\documentclass[10pt,journal,compsoc]{IEEEtran}
\ifCLASSOPTIONcompsoc
  \usepackage[nocompress]{cite}
\else
  \usepackage{cite}
\fi
\ifCLASSINFOpdf
\else
\fi

\usepackage{subcaption,float}
\usepackage[pdftex]{graphicx}
\usepackage{color, colortbl}
\usepackage{amsmath, amsthm}
\usepackage{amssymb}
\usepackage{multirow, array}
\usepackage{bm,url}
\usepackage{breakurl}
\usepackage{enumerate}
\usepackage[table,xcdraw]{xcolor}
\usepackage{hhline}
\usepackage{rotating}
\usepackage{dblfloatfix}
\usepackage{adjustbox}
\usepackage{tikz}
\usepackage{collcell}
\usepackage[ruled,vlined]{algorithm2e}

\usepackage{subcaption}
\usepackage{pgfplots}
\usepackage{hyperref}

\renewcommand{\baselinestretch}{1.0}

\SetKwInput{KwInput}{Input}  % Set the Input
\SetKwInput{KwOutput}{Output} % set the Output

\newtheorem{theorem}{Theorem}
\newtheorem{lemma}{Lemma}
\newtheorem{remark}{Remark}
\newtheorem{hypothesis}{Hypothesis}

\begin{document}

\title{Multiple-Input Variational Auto-Encoder for Anomaly Detection in Heterogeneous Data}

\author{
	\IEEEauthorblockN{
		Phai~Vu~Dinh,
		Diep~N.~Nguyen, ~\IEEEmembership{Senior Member,~IEEE,}
		Dinh~Thai~Hoang,  ~\IEEEmembership{Senior Member,~IEEE,}
		Quang~Uy~Nguyen,
		%Son~Pham~Bao,
		and Eryk~Dutkiewicz}, ~\IEEEmembership{Senior Member,~IEEE}
	
	\thanks{
		  Phai V. D.,  Diep N. Nguyen, D. T. Hoang, and E. Dutkiewicz are with the School of Electrical and Data Engineering, the University of Technology Sydney, Sydney, NSW 2007, Australia (e-mail:   Phai.D.Vu@student.uts.edu.au, \{diep.nguyen, hoang.dinh, eryk.dutkiewicz\}@uts.edu.au). 
		
		N. Q. Uy is with the Computer Science Department, Institute of Information and Communication Technology, Le Quy Don Technical University, Hanoi, Vietnam (e-mail: quanguyhn@lqdtu.edu.vn). 

  		Corresponding author: Phai V. D. (email: Phai.D.Vu@student.uts.edu.au)
		
		%P. B. Son is with the University of Engineering and Technology, Vietnam National University, Hanoi, Vietnam (e-mail: sonpb@vnu.edu.vn).
	}
	\thanks{		
		Preliminary results of this work have been presented at the Global Communications Conference in December 2024, Cape Town, South Africa \cite{dinh2024MIAEAD}.
	}
}

\markboth{IEEE TRANSACTIONS ON PATTERN ANALYSIS AND MACHINE INTELLIGENCE}%
{Shell \MakeLowercase{\textit{et al.}}: Bare Demo of IEEEtran.cls for Computer Society Journals}
% The only time the second header will appear is for the odd numbered pages
% after the title page when using the twoside option.
% 
% *** Note that you probably will NOT want to include the author's ***
% *** name in the headers of peer review papers.                   ***
% You can use \ifCLASSOPTIONpeerreview for conditional compilation here if
% you desire.

% The publisher's ID mark at the bottom of the page is less important with
% Computer Society journal papers as those publications place the marks
% outside of the main text columns and, therefore, unlike regular IEEE
% journals, the available text space is not reduced by their presence.
% If you want to put a publisher's ID mark on the page you can do it like
% this:
%\IEEEpubid{0000--0000/00\$00.00~\copyright~2015 IEEE}
% or like this to get the Computer Society new two part style.
%\IEEEpubid{\makebox[\columnwidth]{\hfill 0000--0000/00/\$00.00~\copyright~2015 IEEE}%
%\hspace{\columnsep}\makebox[\columnwidth]{Published by the IEEE Computer Society\hfill}}
% Remember, if you use this you must call \IEEEpubidadjcol in the second
% column for its text to clear the IEEEpubid mark (Computer Society jorunal
% papers don't need this extra clearance.)

% use for special paper notices
%\IEEEspecialpapernotice{(Invited Paper)}

% for Computer Society papers, we must declare the abstract and index terms
% PRIOR to the title within the \IEEEtitleabstractindextext IEEEtran
% command as these need to go into the title area created by \maketitle.
% As a general rule, do not put math, special symbols or citations
% in the abstract or keywords.
\IEEEtitleabstractindextext{%
\begin{abstract}
Outlier/anomaly detection plays a pivotal role in AI applications, e.g., in classification, and intrusion/threat detection in cybersecurity. However,  most existing methods face challenges of heterogeneity amongst feature subsets posed by non-independent and identically distributed (non-IID) data. To address this, we propose a novel neural network model called Multiple-Input Auto-Encoder for Anomaly Detection (MIAEAD). 
MIAEAD assigns an anomaly score to each feature subset of a data sample to indicate its likelihood of being an anomaly. This is done by using the reconstruction error of its sub-encoder as the anomaly score. All sub-encoders are then simultaneously trained using unsupervised learning to determine the anomaly scores of feature subsets. The final Area Under the ROC Curve (AUC) of MIAEAD is calculated for each sub-dataset, and the maximum AUC obtained among the sub-datasets is selected.
To leverage the modelling of the distribution of normal data to identify anomalies of the generative models, we then develop a novel neural network architecture/model called Multiple-Input Variational Auto-Encoder (MIVAE). MIVAE can process feature subsets through its sub-encoders before learning the distribution of normal data in the latent space. This allows MIVAE to identify anomalies that significantly deviate from the learned distribution.
We theoretically prove that the difference in the average anomaly score between normal samples and anomalies obtained by the proposed MIVAE is greater than that of the Variational Auto-Encoder (VAEAD), resulting in a higher AUC for MIVAE.
Extensive experiments on eight real-world anomaly datasets from different domains, e.g., health, finance, cybersecurity, and satellite imaging, demonstrate the superior performance of MIAEAD and MIVAE over conventional methods and the state-of-the-art unsupervised models, by up to $6\%$ in terms of AUC score.
Furthermore, experimental results show that the AUC obtained by MIAEAD and MIVAE is mostly not impacted as the ratio of anomalies to normal samples in the dataset increases. Alternatively, MIAEAD and MIVAE have a high AUC when applied to feature subsets with low heterogeneity based on the coefficient of variation (CV) score.
\end{abstract}

% Note that keywords are not normally used for peerreview papers.
}

% make the title area
\maketitle

% To allow for easy dual compilation without having to reenter the
% abstract/keywords data, the \IEEEtitleabstractindextext text will
% not be used in maketitle, but will appear (i.e., to be "transported")
% here as \IEEEdisplaynontitleabstractindextext when the compsoc 
% or transmag modes are not selected <OR> if conference mode is selected 
% - because all conference papers position the abstract like regular
% papers do.
\IEEEdisplaynontitleabstractindextext
% \IEEEdisplaynontitleabstractindextext has no effect when using
% compsoc or transmag under a non-conference mode.

% For peer review papers, you can put extra information on the cover
% page as needed:
% \ifCLASSOPTIONpeerreview
% \begin{center} \bfseries EDICS Category: 3-BBND \end{center}
% \fi
%
% For peerreview papers, this IEEEtran command inserts a page break and
% creates the second title. It will be ignored for other modes.
\IEEEpeerreviewmaketitle

\IEEEraisesectionheading{\section{Introduction}
\label{sec:introduction}}

\begin{table}[t] 
	\centering
	\begin{tabular}{c}
		\begin{subfigure}{0.5\textwidth} 
			%\centering
            \hspace{-0.5cm}
			\includegraphics[width=1.0\linewidth, height=0.35\linewidth]{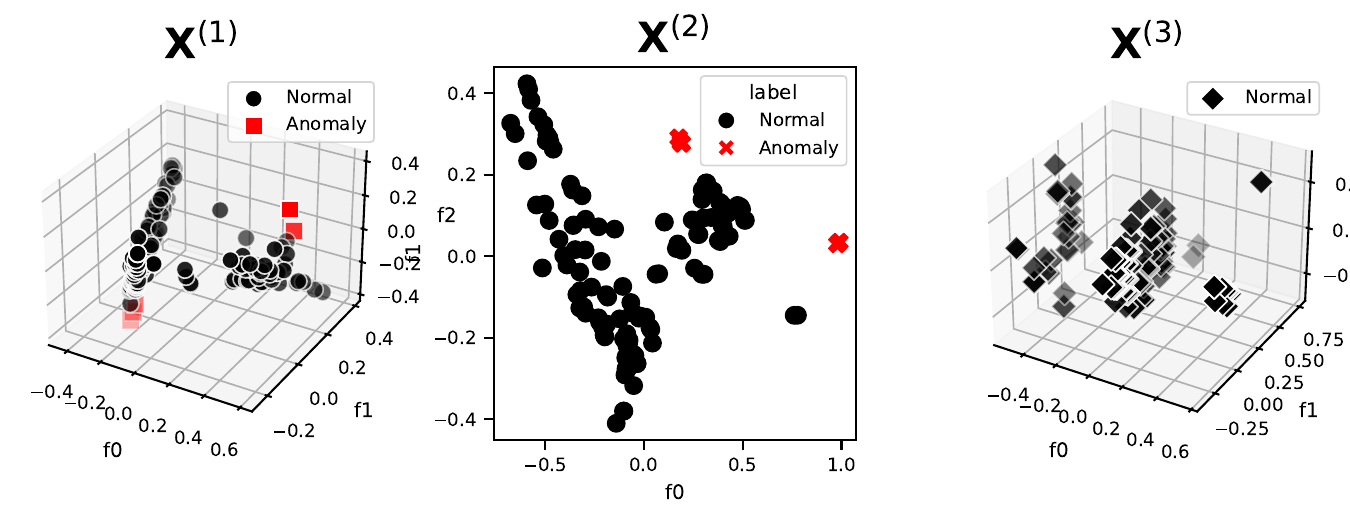} 
			\caption{Data simulation.} 
		\end{subfigure}  \\
		\begin{subfigure}{0.5\textwidth}
			\centering
			\includegraphics[width=0.75\linewidth, height=0.35\linewidth]{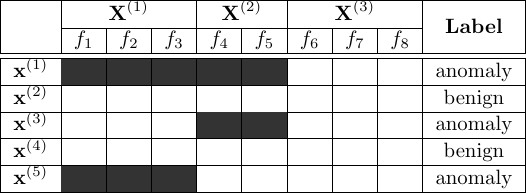}
			\caption{Feature simulation.}
			\vspace{1.5ex} 
		\end{subfigure} \\ \vspace*{-0.5 cm}
		
	\end{tabular}
	\captionof{figure}{An example of a non-IID dataset collected from three different sources. Abnormal features are in black while normal ones are white.}
	
	\label{fig:sub-datasets}
\end{table}

\IEEEPARstart{O}{utlier/anomaly} detection plays an essential role in many applications, i.e., network security, medical diagnosis, and fraud detection \cite{9271895, ahmed2021graph, luo2021future, 10636792, 10273635, xiang2024exploiting}. However, for non-independent and identically distributed (non-IID) data, detection methods face critical challenges by the coupling between data samples and the heterogeneity amongst feature subsets, deteriorating the accuracy of the detection engine \cite{cao2014non, herlands2018gaussian}. For non-IID data, the coupling problem likely exists between data samples or amongst features of a data sample \cite{pang2021homophily, pang2017learning}. For example, when fraudsters steal a credit card, they may attempt multiple purchases before the card is locked. Consequently, these transactions are not independent because they are linked to fraudulent activity that started after the card was stolen. Additionally, the heterogeneous data presents differences in their distributions, heterogeneity of feature subsets, and non-identical distributions of data subsets \cite{cao2014non}. For instance, IoT intrusion detection systems collect data from various resources, e.g., network traffic, log systems, and webpage content, resulting in the final feature space comprised of heterogeneous feature subsets. Therefore, anomaly features are more likely to be present within a feature subset rather than across all features of a sample.
As illustrated in Fig. \ref{fig:sub-datasets}, a non-IID dataset $\mathbf{X}$ consists of three sub-datasets, namely $\mathbf{X}^{(1)}$, $\mathbf{X}^{(2)}$, and $\mathbf{X}^{(3)}$, collected from three sources, i.e., network traffic, log systems, and webpage content, respectively. 
As observed in Fig. \ref{fig:sub-datasets} (a), anomalies may be distinguished from the normal samples in the sub-datasets, i.e., $\mathbf{X}^{(1)}$ with three-dimensional space and $\mathbf{X}^{(3)}$ with two-dimensional space, whilst there are only normal samples in the sub-dataset $\mathbf{X}^{(3)}$ with three-dimensional space.
Fig. \ref{fig:sub-datasets} (b) presents feature sets $(f_1, f_2, f_3)$, $(f_4, f_5)$, and $(f_6, f_7, f_8)$  which correspond to $\mathbf{X}^{(1)}$, $\mathbf{X}^{(2)}$, and $\mathbf{X}^{(3)}$, respectively. $f_1$, $f_2$, and $f_3$ may be extracted from the information of protocol type, service, and flag of the packets, whilst $f_4$, $f_5$, $f_6$, $f_7$, and $f_8$, are hidden/implicit features extracted from log systems and webpages' content. The feature set $(f_1, f_2, f_3, f_4, f_5, f_6, f_7, f_8)$ represents $\mathbf{X}$.

Without assuming any specific data distribution, conventional non-parametric anomaly detection methods based on density function, e.g., local outlier factor (LOF) and kernel density estimation (KDE) \cite{tang2017local},  and one-class support vector machine (OSVM) \cite{OSVM} are often adopted, especially to deal with the heterogeneity of data. Specifically, LOF calculates the local density of each data sample to its neighbours, whilst KDE estimates the overall density of the dataset. The data samples with density scores lower than a threshold are identified as anomalies.
OSVM finds a hyperplane to separate normal data samples from the origin \cite{OSVM}. Samples on the minority side of the hyperplane are considered anomalies. However, all these above methods are sensitive to parameter tuning, as non-parametric methods tend to be. 
Additionally, these methods may not capture non-linear relationships between data samples and amongst feature subsets, leading to lower accuracy when applied to high-dimensional non-IID data.

Thanks to its ability to capture non-linear relationships amongst feature subsets and deal well with high-dimensional data, Auto Encoder (AE) based architectures have recently emerged as potential solutions for detecting anomalies, especially for high-dimensional non-IID datasets. More importantly, the AE can extract the most important hidden/latent features of the input data at its bottleneck layer  \cite{pang2021deep}. The AE-based methods assign an anomaly score (capturing the likelihood of being an anomaly \cite{RandNet, RDAE, DAGMM}) to a data sample based on its reconstruction error, which is calculated from all feature subsets.  Since the AE prioritizes reconstructing the output as close to its input from the major normal samples as possible, data with anomalies typically have higher anomaly scores compared to benign samples. The authors of \cite{RDAE} propose Robust Deep AE (RDA), whilst RandNet \cite{RandNet} is another approach that randomly removes connections amongst layers of the AE to create multiple models. The authors of \cite{DAGMM} combine data from the latent space with the reconstruction error value of the AE, and input this combination into a Gaussian Mixture Model (DAGMM) to predict the likelihood of being as an anomaly or a benign sample. RandNet and DAGMM are considered state-of-the-art unsupervised anomaly detection models \cite{RandNet, DAGMM, AEWAD}. However, most AE-based methods, as aforementioned, use the entire feature set to calculate the anomaly score. In practice,  due to the heterogeneity of non-IID data, as observed in Fig. \ref{fig:sub-datasets}, anomaly features are more likely to be present within a feature subset rather than across the entire features of a sample. Additionally, methods like RDA are more sensitive to hyper-parameters, i.e., the ratio of anomalies identified/removed from the training set \cite{RDAE} after each training epoch. They, e.g., RandNet, also require high computing costs due to training many separate AE models. The trade-off between the reconstruction and regularized terms during the training process of DAGMM may result in lower accuracy for anomaly detection \cite{DAGMM}.

Instead of using the entire feature set, one can rely on only subsets of these features. For example, the authors of \cite{pang2017learning} propose a wrapper-based outlier detection framework (WrapperOD), which optimizes feature subset selection and ranks data samples based on a given feature subset. Another method, outlier detection embedded feature selection (ODEFS) was proposed in \cite{cheng2020outlier} to learn feature subsets specifically for outlier detection. Although WrapperOD and ODEFS show promising results in anomaly detection, they face challenges in dealing with high-dimensional feature spaces, requiring excessively high computing costs. Moreover, as these methods, including \cite{PerFedMask}, separately learn different feature subsets based on federated learning, they may discard important information from non-selected features of different data sources. To deal with multiple data sources, the authors in \cite{MIAE-ICC, dinh2024multipleinput} introduce Multiple-Input Auto-Encoder (MIAE) which uses multiple sub-encoders to process data from multiple sources. The representation data extracted from the bottleneck layer of MIAE is then used for classifiers in the intrusion detection systems.
However, the methods in \cite{PerFedMask, MIAE-ICC, dinh2024multipleinput} were not thoroughly explored the anomaly detection problem.

To leverage the latent space to enhance the AUC of anomaly detection methods, the authors in \cite{an2015variational} proposed Variational AE for Anomaly Detection (VAEAD), which aims to model the distribution of normal data in the latent space to identify anomalies. Abnormal samples generated from the latent space of the VAEAD may significantly deviate from the distribution of normal data. As a result, the decoder of VAEAD struggles to reconstruct abnormal samples at its output, leading to higher reconstruction errors for anomalies compared to normal samples.
The Generative Adversarial Network (AnoGAN) \cite{AnoGAN}, fast AnoGAN \cite{f-AnoGAN}, and skip AnoGAN \cite{SkipAnoGAN} were introduced for anomaly detection. Their generators aim to generate fake normal samples, while the discriminator distinguishes between real normal samples and fake samples. These methods combine both the generator’s reconstruction error and the discriminator’s reconstruction error in the latent space to identify anomalies.
The authors in \cite{AAE} introduced Adversarial AE (AAE), which leverages both the reconstruction error of the AE and incorporates a discriminator in the latent space of the AE to distinguish between real and fake data samples in the latent space. AAE also combines both reconstruction errors in the input space and the latent space to identify anomalies. However, the above generative models may struggle to identify anomalies that are present in feature subsets rather than in the entire feature space.

Given the above, we first propose a novel neural network model called Multiple-Input Auto-Encoder for Anomaly Detection (MIAEAD) that can deal effectively with heterogeneous/non-IID and high dimensional data. This is achieved by leveraging the data from different sources, capturing the non-linear relationship among these heterogeneous samples/sources while requiring only feature subsets. To this end, MIAEAD assigns an anomaly score to each feature subset which is the reconstruction error of its sub-encoder. All sub-encoders of MIAEAD are then simultaneously trained in an unsupervised learning manner to determine the anomaly scores of feature subsets. The final Area Under the ROC Curve (AUC) of MIAEAD is calculated for each sub-dataset, and the maximum AUC obtained among the sub-datasets is selected. %However, MIAEAD does not model the distribution of normal data in the latent space, which limits its ability to fully leverage the latent space structure for anomaly detection.
To leverage the modelling of the normal data distribution to identify anomalies in MIAEAD, we then propose a novel neural network architecture/model called Multiple-Input Variational Auto-Encoder (MIVAE). MIVAE can process feature subsets through its sub-encoders before learning the distribution of normal data in the latent space. This allows MIVAE to identify anomalies that significantly deviate from the learned distribution. As a result, the reconstruction error of the feature subsets for anomalies may be significantly greater than that of normal samples.
We conduct extensive experiments on eight anomaly datasets from different domains (health, finance, cybersecurity, satellite imaging, etc) including cardio disease (M1), credit card fraud (M2) from Kaggle, a non-IID dataset called Arrhythmia (M3),  medical image Mammography (M4), network security NSLKDD (M5),  satellite image (M6), Shuttle (M7), and Spambase (M8) from the UCI repository \cite{AEWAD}. The results show that MIAEAD and MIVAE outperform conventional anomaly methods and the state-of-the-art unsupervised models, i.e., RandNet \cite{RandNet} and DAGMM \cite{DAGMM}, and the generative models, i.e., AnoGAN \cite{AnoGAN}, SkipGAN \cite{SkipAnoGAN}, and AAE \cite{AAE}. The major contributions are summarized as follows:	
\begin{itemize}
	\item We first propose a novel neural network model, i.e., MIAEAD, which detects anomalies by using feature subsets instead of all features of a non-IID data sample. MIAEAD is trained using an unsupervised learning method, processing all feature subsets simultaneously to precisely detect anomalies in each subset. 
	\item We then develop a novel neural network architecture/model, called MIVAE. MIVAE can process feature subsets through its sub-encoders before learning the distribution of normal data in the latent space. This allows MIVAE to identify anomalies that significantly deviate from the learned distribution. 
	\item We theoretically prove that the difference in the average anomaly score between normal samples and anomalies obtained by the proposed MIVAE is greater than that of the Variational Auto-Encoder (VAEAD), resulting in a higher AUC for MIVAE. The MIAEAD and MIVAE models also use fewer parameters than the AEAD and VAEAD models.  
	\item We conduct extensive experiments to evaluate the performance of MIAEAD and MIVAE compared to other methods on eight benchmark anomaly datasets from different domains. The Area Under the ROC Curve (AUC) achieved by MIAEAD and MIVAE is significantly higher than that of the state-of-the-art anomaly detection methods, namely RandNet \cite{RandNet} and DAGMM \cite{DAGMM}, by up to $6\%$. Moreover, MIAEAD and MIVAE exhibit an average Miss Detection Rate (MDR) of $12.5\%$ and $11.9\%$ across all eight datasets. The AUC obtained by MIAEAD and MIVAE is relatively steady (i.e., not deteriorating much) as the ratio of anomalies to normal samples in the dataset increases. We observe that MIAEAD and MIVAE have a high AUC when applied to feature subsets with low heterogeneity based on the coefficient of variation (CV) score \cite{verrill2007confidence}. 
	%\item We extensively analyze the characteristics of the MIAEAD and MIVAE models to further demonstrate their superior performance over other methods in identifying anomalies using the entire feature space. 
\end{itemize}

%\section{related work}

\begin{table}[t] %\vspace{0.02}
	\caption{Variables and Notations used in the paper.}
	\label{tab:variable_name}
	\setlength\tabcolsep{0.0pt}
	\centering
	\small
	%\scriptsize
	%\footnotesize	
	\begin{tabular}{l | l}
		\hline
		$\mathbf{X}$                        & Dataset $\mathbf{X}$                                    \\ %\hline
		$N$                        & Number of samples of $\mathbf{X}$                       \\ %\hline
		$\mathbf{x}^{(i)}$ & The $i^{th}$ data sample of $\mathbf{X}$                \\ %\hline
		$d$                        & Dimensionality of $\mathbf{x}^{(i)}$ \\ %\hline
		$n_a$                        & Number of anomalies of $\mathbf{X}$ \\ %\hline
		$n_b$                        & Number of benign samples of $\mathbf{X}$ \\ %\hline
		$s_{(.)}(\mathbf{x}^{(i)})$                        & Anomaly score of data sample $\mathbf{x}^{(i)}$ \\ %\hline
		$(.)$                        & Method is used \\ %\hline
		$d_z$                        & Dimensionality of the latent space $\mathbf{z}$ \\ %\hline
		$L$                        & Number of sampling $\mathbf{z}$ \\ %\hline
		$M$                        & Number of sub-datasets \\ %\hline
		$\mathbf{X}^{(j)}$                        & The $j^{th}$ sub-dataset  \\ %\hline
		$\mathbf{x}^{(i,j)}$                        & The $i^{th}$ data sample of the sub-dataset $\mathbf{X}^{(j)}$ \\ %\hline
		$d_j$                        & The dimensionality of the sub-dataset $\mathbf{X}^{(j)}$ \\ %\hline
		$\mathbf{e}^{(i,j)}$                        & The output of the sub-encoder $j^{th}$  \\ %\hline
		$\boldsymbol{\mu}^{(i)}$                        & The mean of the learned latent variable\\ %\hline
		$\boldsymbol{\sigma}^{(i)}$                        & The standard deviation \\ %\hline
		$\boldsymbol{\epsilon}^{(i,l)}$                        & The random variable sampled \\ %\hline
		$\bigoplus$                        & The combination operator\\ %\hline
		$\boldsymbol{\phi}$                        & The parameter sets of encoder\\ %\hline
		$\boldsymbol{\theta}$                        & The parameter sets of decoder\\ %\hline
		$\mathbf{W}, \mathbf{b}$                        & The weights and biases of the neural network\\ %\hline
		$CV$                        & The coefficient of variation\\ % \hline
		$\beta^{(j)}$                        & The balanced hyper-parameter of the $j^{th}$ branch \\ %\hline
		$\delta$                        & The average anomaly score of normal samples\\ %\hline
		$E$                        & The average anomaly score of anomalies\\ %\hline
		$\Delta s_{(.)}$                        & $E-\delta$\\ %\hline
		$\nu = \frac{E}{\delta}$ & $\nu$ is the ratio of $E$ by $\delta$ \\ %\hline
		$T$                        &The number of layers of the encoder\\ %\hline
        $h_{i}^{(t)}$ & The activation of node $i$ in the layer $t$\\ %\hline
        $\sigma_{k}^{(t)}$ & The error term for node $k$ in layer $t$\\ %\hline
        $\odot$ & The element-wise operation\\ %\hline
        $\nabla W_{ik}^{(t)}$ & The gradients of the weights \\ %\hline
        $\nabla b_{i}^{(t)}$ & The gradients of the biases \\ \hline
		
	\end{tabular}

\end{table}

\section{BACKGROUND}
\subsection{Problem Statement}
\label{l_general_problem_statement}
Given a dataset of $N=n_a + n_b $ samples $\mathbf{X}= \{ \mathbf{x}^{(1)},  \mathbf{x}^{(2)}, \ldots,  \mathbf{x}^{(N)} \}$ with $ \mathbf{x}^{(i)} \in \mathbb{R}^{d}$, $i = \{1, 2, \ldots, N \}$, $d$ is dimensionality of $\mathbf{x}^{(i)}$. $n_a$ and $n_b$ present the numbers of abnormal and benign samples, respectively. The goal is to find $n_a$ abnormal samples from the dataset $\mathbf{X}$. This can be achieved by learning a scoring function $s: \mathbf{X} \to \mathbb{R}$ that assigns scores to data samples, such that $s(\mathbf{x}^{(i)})  <  s(\mathbf{x}^{(k)})$ where $\mathbf{x}^{(i)}$ represents a normal sample and $\mathbf{x}^{(k)}$ represents an abnormal sample. Here, $i \in \{1, 2, \ldots, n_b\}$ and $k \in \{n_b+1, n_b+2, \ldots,n_a + n_b \}$.

\subsection{Auto-Encoder For Anomaly Detection (AEAD)}
\label{l_aead}
%Let $\mbox{s:}$ $ \mathbf{X} \to \mathbb{R}$ be a neural network, i.e., an Auto-Encoder (AE), which is a fundamental method in developing our model in the next section. 
The AE architecture consists of two components, i.e., an Encoder and a Decoder. The Encoder uses the function $f(\mathbf{x}^{(i)}, \boldsymbol{\phi})$ to map the input sample $\mathbf{x}^{(i)}$ into the latent space $\mathbf{z}^{(i)} \in \mathbb{R}^{d_z}$, where $d_z$ represents dimensionality of  $\mathbf{z}^{(i)}$,  $d_z < d$. $\boldsymbol{\phi} = (\mathbf{W}^e, \mathbf{b}^e)$ are weights and biases of the neural network of the Encoder. Next, the Decoder applies the function $g(\mathbf{z}^{(i)}, \boldsymbol{\theta})$ to map the latent sample $\mathbf{z}^{(i)}$ into $\hat{\mathbf{x}}^{(i)} \in \mathbb{R}^d$, where $\boldsymbol{\theta} = (\mathbf{W}^e, \mathbf{b}^e)$ are weights and biases of the Decoder. The AE aims to reconstruct the output $\hat{\mathbf{x}}^{(i)}$ as closely as possile to the input $\mathbf{x}^{(i)}$. Consequently, the anomaly score of the data sample $\mathbf{x}^{(i)}$ is  assigned to the reconstruction error: %$s^{(i)} = s(\mathbf{x}^{(i)})=\frac{1}{d} \sum_{t=1}^{d}\left({x}^{(i)}_{t}-\hat{{x}}^{(i)}_{t}\right)^{2},$
\begin{equation}
	\label{eq:score_ae}
	s_{\emph{AEAD}}(\mathbf{x}^{(i)}) = \frac{1}{d} \sum_{t=1}^{d}\left({x}^{(i)}_{t}-\hat{{x}}^{(i)}_{t}\right)^{2} .
\end{equation} 
The AE-based anomaly detection is trained to minimize the loss function: $\ell_{\emph{AEAD}}(\mathbf{X}, \boldsymbol{\phi}, \boldsymbol{\theta})=\frac{1}{N} \sum_{i=1}^{N} ( s^{(i)}), $ 
%	\begin{equation}
	%		\label{eq:ae_loss_mse}
	%		\ell_{\mbox{AE}}(\mathbf{X}, \boldsymbol{\phi}, \boldsymbol{\theta})=\frac{1}{N} \sum_{i=1}^{N}\left( s^{(i)}
	%		 \right),
	%	\end{equation}
where $N$ is the number of samples of the dataset $\mathbf{X}$. After training the AE model, we select the top $n_a$ largest values $s^{(i)}$ from the list  $(s^{(1)}, s^{(2)}, \ldots, s^{(N)}) $. From the selected $s^{(i)}$ , we can find $n_a$ data samples $\mathbf{x}^{(i)}$ to assign as abnormal labels. One might question why the largest values $s^{(i)}$ are considered anomalies. This is because the number of anomalies is often significantly lower than the number of normal samples ($n_a << n_b$). As a result, the AE tends to have a lower reconstruction error (anomaly score) for the normal samples compared to the abnormal ones. In addition, if $n_a \approx n_b$ or  $n_a >> n_b$, one might question the accuracy of AE-based anomaly detection models. This will be discussed in more detail in subsection \ref{label:influence_ratio_anomaly_benign}.

\subsection{Variational Auto-Encoder For Anomaly Detection (VAEAD)}
To leverage a probabilistic latent space for capturing the data distribution of normal data, the authors in \cite{an2015variational} used a Variational Autoencoder for Anomaly Detection (VAEAD). 
The  Evidence Lower Bound (ELBO) for a single data sample $\mathbf{x}^{(i)} \in \mathbf{X} = \{ \mathbf{x}^{(i)} \}_{i=1}^{N} $ is as follows:
\begin{equation}
	\label{eq:elbo-vae}
	\begin{aligned}
		\mathcal{L}_{\emph{VAE}}^{(i)} =     \mathbb{E}_{q(\mathbf{z} | \mathbf{x}^{(i)})} \left[ \log p(\mathbf{x}^{(i)} | \mathbf{z}) \right] - D_{\emph{KL}} \left( q(\mathbf{z} | \mathbf{x}^{(i)}) \| p(\mathbf{z}) \right),
	\end{aligned}
\end{equation}
where the first term in Eq. (\ref{eq:elbo-vae}) is the expected log-likelihood of reconstructing the input $\mathbf{x}^{(i)}$ from the latent space $\mathbf{z}$, whilst the second term measures the KL-divergence between the posterior distribution $q(\mathbf{z} | \mathbf{x}^{(i)})$ and the prior distribution $p(\mathbf{z})$.

In practice, the encoder of VAEAD maps input data into the latent space as $\mathbf{e}^{(i)} = f(\mathbf{x}^{(i)}, \boldsymbol{\phi})$, where $\boldsymbol{\phi} = (\mathbf{W}^e, \mathbf{b}^e)$ are weights and biases of the neural network. The mean $\boldsymbol{\mu}^{(i)} = f_{\mu}(\mathbf{e}^{(i)})$ and standard deviation $\boldsymbol{\sigma}^{(i)} = f_{\sigma}(\mathbf{e}^{(i)})$ represent the distribution of the input $\mathbf{x}^{(i)}$ in the latent space. Unlike AEAD, the latent space in VAEAD aims to force the data samples to follow a distribution of the normal data. To achieve this, a latent vector $\mathbf{z}^{(i)}$ is sampled from the learned distribution, defined as $\mathbf{z}^{(i, l)} = \boldsymbol{\mu}^{(i)} + \boldsymbol{\sigma}^{(i)}  \boldsymbol{\epsilon}^{(i,l)}$, where $\boldsymbol{\epsilon}^{(i,l)} \sim \mathcal{N}(0, I)$ and $l \in \{1, 2, \ldots, L\}$ represents the $l^{\text{th}}$ sampling iteration. The VAEAD aims to reconstruct the input $\mathbf{x}^{(i)}$ at the output of the decoder, i.e., $\hat{\mathbf{x}}^{(i,l)} = f(\mathbf{z}^{(i,l)}, \boldsymbol{\theta})$, where $\boldsymbol{\theta} = (\mathbf{W}^d, \mathbf{b}^d)$ are weights and biases of the neural network of the decoder. To ensure that the reconstruction $\hat{\mathbf{x}}^{(i,l)}$ is as close to the original input $\mathbf{x}^{(i)}$ as possible, VAEAD uses the KL-divergence to regularize the learned posterior distribution $q(\mathbf{z}|\mathbf{x}^{(i)})$, encouraging it to be close to the prior distribution $p(\mathbf{z}) \sim \mathcal{N}(0, I)$, i.e., $D_{\emph{KL}}(q_{\phi}(\mathbf{z} | \mathbf{x}^{(i)}) \parallel p(\mathbf{z})) = \frac{1}{2} \sum_{t=1}^{d_{z}} \big( 1 + \log ((\sigma_t^{(i)})^2) - (\mu_t^{(i)})^2 - (\sigma_t^{(i)})^2 \big)$, where $d_{z}$ is dimensionality of the latent space $\mathbf{z}^{(i)}$ \cite{kingma2014auto}. The KL-divergence  $D_{\emph{KL}}(q_{\phi}(\mathbf{z} | \mathbf{x}^{(i)}) \parallel p(\mathbf{z}))$ ensures that the stochastic latent variable  $\mathbf{z}$ keep enough information from the original input $\mathbf{x}^{(i)}$ to reconstruct at the output of the decoder. In addition, the generated latent variables $\mathbf{z}^{(i,l)}$ for anomalies are likely to differ from those of the normal data, causing the VAEAD to struggle with reconstructing the abnormal samples at its output. Therefore, the anomaly score of the VAEAD of data sample $\mathbf{x}^{(i)}$ is assigned to the reconstruction error \cite{an2015variational}, as follows:
\begin{equation}
	\label{eq:score_vae}
	s_{\emph{VAEAD}}^{(i)} = \frac{1}{L  d} \sum_{l=1}^{L} \sum_{t=1}^{d}\left({x}^{(i)}_{t}-\hat{{x}}^{(i, l)}_{t}\right)^{2} .
\end{equation}
Note that VAEAD is trained using a loss function that sums the KL-divergence and the reconstruction error. In addition, VAEAD is generally considered more effective than AEAD for anomaly detection, as the KL-divergence term in VAEAD helps distinguish anomalies from the learned distribution of normal data. When anomalies deviate from the normal data distribution, VAEAD struggles to reconstruct their input, leading to higher reconstruction errors.

%%%%%%%%%%%%%%%%%%%%%%%%%%%%%%%%%%%%%%%%%%%%%%%%%%%%%%%%%%%%%%%%%%%%%%%%%%%%%%%%%%%%%
\section{THE PROPOSED APPROACH}
\subsection{Equivalent Problem Statement}

Given a general problem described in \ref{l_general_problem_statement}, we aim to identify anomalies in a non-IID dataset by using feature subsets instead of all features. Let $\mathbf{X} = \{\mathbf{X}^{(1)}, \mathbf{X}^{(2)}, \ldots \mathbf{X}^{(M)} \}$ be  a non-IID dataset, where $\mathbf{X}^{(j)}= \{\mathbf{x}^{(1, j)}, \ldots, \mathbf{x}^{(N, j)} \}$ is a sub-dataset with dimensionality $d_j$, and $j = \{1, \ldots, M\}$. Each $\mathbf{x}^{(i,j)} \in \mathbb{R}^{d_j} $ represents the $i^{th}$ data sample of the $j^{th}$ sub-dataset $\mathbf{X}^{(j)}$, and $N$ is the number of data samples in $\mathbf{X}^{(j)}$, where $ 1 \leq M \leq d$. $F^{(j)} = \{ f_{(j,1)}, \ldots, f_{(j,d_j)} \}$ denotes the set of features of sub-dataset $\mathbf{X}^{(j)}$. Note that, all sub-datasets $\mathbf{X}^{(j)}$ have the same size, which is $N$. In cases where the number of samples in the sub-datasets differs, it is possible to use undersampling techniques to achieve a balance among these sub-datasets \cite{bagui2021resampling}.
Let's have $\mathbf{X} = \{ \mathbf{x}^{(1)},  \mathbf{x}^{(2)}, \ldots,  \mathbf{x}^{(N)} \}$, where $\mathbf{x}^{(i)} \in \mathbb{R}^{d}$ is the $i^{th}$ data sample of $\mathbf{X}$. Here $d=\sum_{j=1}^{M}(d_j)$ as the dimensionality of $\mathbf{x}^{(i)}$, and $F = \{f_{1}, \ldots, f_{d}\} = F^{(1)} \cup F^{(2)} \cup \ldots \cup F^{(M)} $ as a feature set of $\mathbf{X}$. $\mathbf{x}^{(i)} = \mathbf{x}^{(i,1)}  \bigoplus \ldots \bigoplus \mathbf{x}^{(i,M)}$, where $\bigoplus$ denotes the combination operator. The combination operator is implemented by concatenating $M$ vectors, i.e., $\mathbf{x}^{(i,j)}$, $j=\{1, \ldots, M\}$. For example, if $\mathbf{x}^{(i,1)}=\{1,2\}, \mathbf{x}^{(i,2)}=\{3,4\}$, and $M=2$ then $\mathbf{x}^{(i)} = \{1, 2, 3, 4 \}$.
The goal of finding $n_a$ abnormal samples from the $N$ samples of dataset $\mathbf{X}$ in subsection \ref{l_general_problem_statement} is equivalent to simultaneously finding $n_a$ abnormal samples from all sub-datasets $\{\mathbf{X}^{(1)}, \mathbf{X}^{(2)}, \ldots, \mathbf{X}^{(M)} \}$. One can apply anomaly detection methods/models on each sub-dataset $\mathbf{X}^{(j)}$ and then use ensemble techniques to find the final anomaly detection model \cite{RandNet}.

%%%%%%%%%%%%%%%%%%%%%%%%%%%%%%%%%%%%%%%%%%%%%%%%%%%%%%
\subsection{The Proposed Multiple-Input Auto-Encoder For Anomaly Detection (MIAEAD)}
\begin{figure}[t]
	\vspace*{0ex}
	\centering
	\includegraphics[width=0.40\textwidth]
	{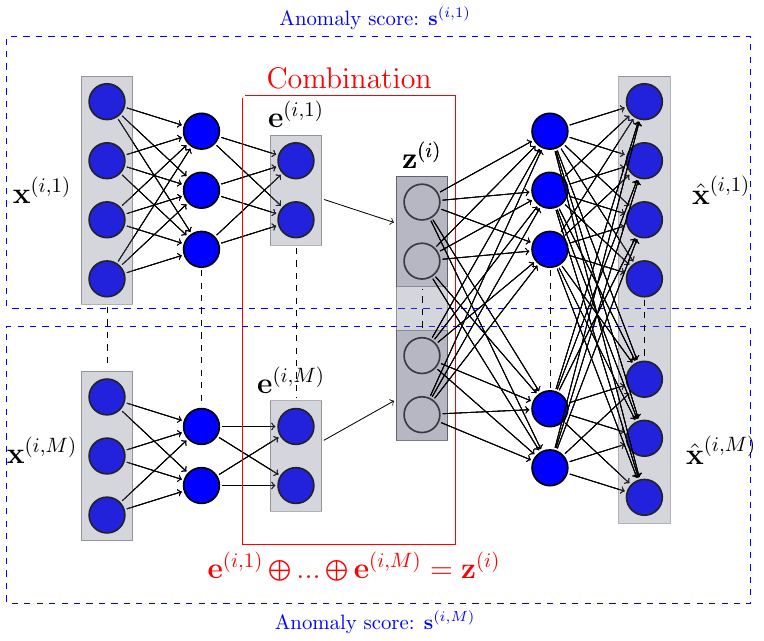}
	\caption{MIAEAD architecture.}
	\label{fig:miae_architecture} 
\end{figure}

To learn an anomaly scoring function $\mbox{s:}$ $\mathbf{X} \to \mathbb{R}$ based on AE in subsection \ref{l_aead} with sub-datasets $\{\mathbf{X}^{(1)}, \mathbf{X}^{(2)}, \ldots \mathbf{X}^{(M)} \}$, we need to create $M$ separate AE models. This can lead to the problem of learning many models with inconsistent goals. To tackle this,
the MIAE model was first presented in \cite{dinh2024multipleinput} to transfer the heterogeneous input with different dimensionality into a lower-dimensional space, which facilitates classifiers. The representation data of MIAE are extracted from the bottleneck layer before being fed to classifiers. Note that, MIAE is a single model with only a loss function. Based on the MIAE architecture, we propose a novel anomaly detection model referred to as MIAEAD, as illustrated in Fig. \ref{fig:miae_architecture}.
First, sub-data samples $\mathbf{x}^{(i,1)}, \ldots, \mathbf{x}^{(i, M)}$ are simultaneously put into sub-encoders of MIAEAD before being transferred into a lower-dimensional space at the output of the sub-encoders, i.e., $\mathbf{e}^{(i,1)}, \ldots, \mathbf{e}^{(i,M)}$. Subsequently, $\mathbf{z}^{(i)}$ is calculated as: $\mathbf{z}^{(i)} = \mathbf{e}^{(i,1)}  \bigoplus \ldots \bigoplus \mathbf{e}^{(i,M)}$.
%\begin{equation}
%	\label{eq:zi}
%	\mathbf{z}^{(i)} = \mathbf{e}^{(i,1)}  \bigoplus \ldots \bigoplus \mathbf{e}^{(i,M)}.
%\end{equation}
Next, $\mathbf{z}^{(i)}$ is put into the Decoder to obtain the $\hat{ \mathbf{x}}^{(i)}$ at its output. Because the numbers of neurons of the Encoder and Decoder are equal, we can separate $\hat{ \mathbf{x}}^{(i)}$ into $(\hat{\mathbf{x}}^{(i,1)}, \ldots, \hat{\mathbf{x}}^{(i, M)} )$ where:
%\begin{equation}
%	\label{eq:xhat_i}
$\hat{\mathbf{x}}^{(i)} = \hat{\mathbf{x}}^{(i,1)}  \bigoplus \ldots \bigoplus \hat{\mathbf{x}}^{(i,M)}$.
%\end{equation}
Here, dimensionality of $\mathbf{x}^{(i,j)}$ and $\hat{\mathbf{x}}^{(i,j)}$ is equal to $d_j$. In this way, we can assign an anomaly score of a sub-data sample $\mathbf{x}^{(i,j)}$ as the reconstruction error of the $j^{th}$ sub-encoder as:
\begin{equation}
	\label{eq:score_miae_ij}
	s_{\emph{MIAEAD}}(\mathbf{x}^{(i,j)})=\frac{1}{d_j} \sum_{t=1}^{d_j}\left({x}^{(i,j)}_{t}-\hat{{x}}^{(i,j)}_{t}\right)^{2}.
\end{equation} 
The anomaly score of the sub-dataset $\mathbf{X}^{(j)}$ is measures as:
\begin{equation}
	\label{eq:score_miae_Xj}
	s({\mathbf{X}^{(j)}}) =  \frac{1}{N  d_j }  \sum_{i=1}^{N}  \sum_{t=1}^{d_j}   \left({x}^{(i,j)}_{t}-\hat{{x}}^{(i,j)}_{t}\right)^{2}.
\end{equation} 
The anomaly score of all sub-datasets, i.e., $\mathbf{X}^{(1)}, \mathbf{X}^{(2)}, \ldots, \mathbf{X}^{(M)}$, is measures as:
\begin{equation}
	\label{eq:score_miae_X}
	s({\mathbf{X}}) = \frac{1}{N}  \sum_{j=1}^{M} \bigg(\frac{1}{d_j} \sum_{i=1}^{N}  \sum_{t=1}^{d_j}   \left({x}^{(i,j)}_{t}-\hat{{x}}^{(i,j)}_{t}\right)^{2} \bigg).
\end{equation} 
After training the MIAEAD with a loss function $s({\mathbf{X}})$, $s^{(i, j)}$ is measured to determine $\mathbf{x}^{(i, j)}$ as anomaly or benign sample in the sub-dataset $\mathbf{X}^{(j)}$. If $\mathbf{x}^{(i, j)}$ is assigned as anomaly, data sample $\mathbf{x}^{(i)}$ is assigned as anomaly in the final dataset $\mathbf{X}$. Note that, unlike MIAE which reconstructs the input $\mathbf{x}^{(i)}$ at its output $\hat{\mathbf{x}}^{(i)}$, MIAEAD tries to separately reconstruct sub-data sample $\mathbf{x}^{(i,j)}$ at its output $\hat{\mathbf{x}}^{(i,j)}$. Therefore, MIAEAD can detect anomalies by the sub-data sample $\mathbf{x}^{(i,j)}$ instead of using the data sample $\mathbf{x}^{(i)}$.
\begin{figure}[t]
	\vspace*{0ex}
	\centering
	\includegraphics[width=0.44\textwidth]
	{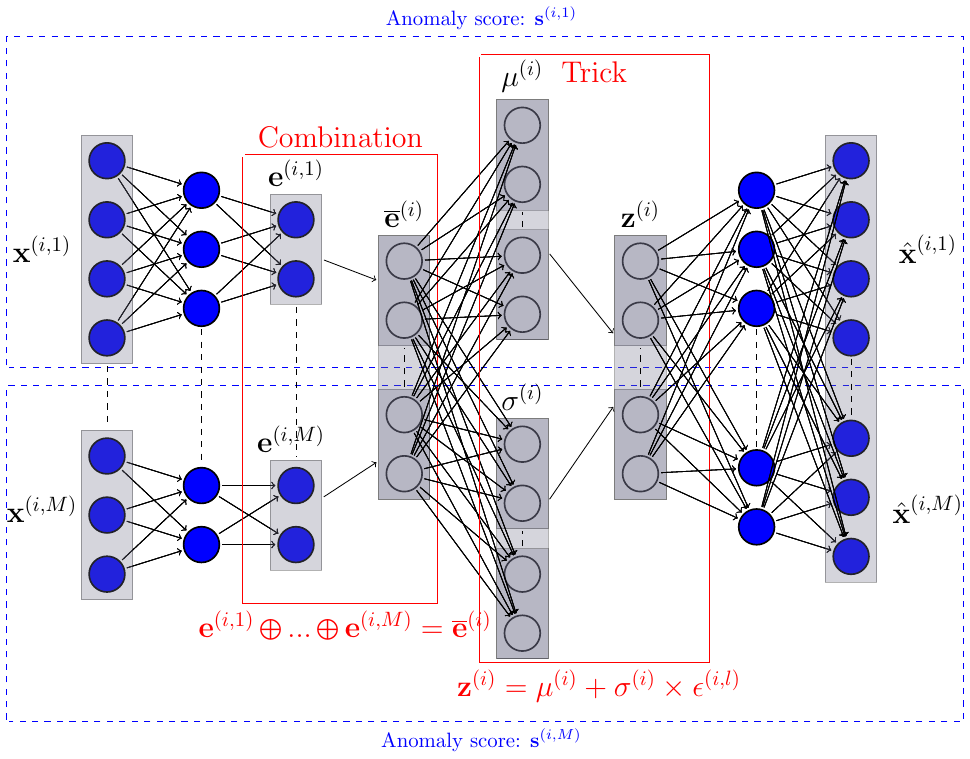}
	\caption{The proposed MIVAE architecture. MIVAE consists of multiple sub-encoders that simultaneously process data, including different feature subsets.}
	\label{fig:mivae_architecture} 
\end{figure}
\subsection{The Proposed Multiple-Input Variational Auto-Encoder For Anomaly Detection (MIVAE)}
To leverage the effectiveness of MIAEAD in assigning anomaly scores to each feature subset of a data sample and distinguishing anomalies from the learned distribution of normal data in the latent space of VAEAD, we propose the Multiple-Input Variational Autoencoder for Anomaly Detection (MIVAE) architecture/model, as illustrated in Fig. \ref{fig:mivae_architecture}.
Unlike the original VAE model, the Evidence Lower Bound (ELBO) for a single data sample $\mathbf{x}^{(i)} = \mathbf{x}^{(i,1)}  \bigoplus \ldots \bigoplus \mathbf{x}^{(i, M)}$ of MIVAE is formulated as follows:
\begin{equation}
	\label{eq:elbo-mivae}
	\begin{aligned}
		\mathcal{L}_{\emph{MIVAE}}^{(i)} =  \frac{1}{L}\sum_{l=1}^{L}\sum_{j=1}^{M} \bigg[  \mathbb{E}_{q(\mathbf{z}^{(i,l)} | \mathbf{x}^{(i,j)})} \bigg[ \sum_{t=1}^{d_j} \log p({x}_t^{(i,j)} | \mathbf{z}^{(i,l)}) \bigg]  \\
		-  D_{\emph{KL}}\left( q(\mathbf{z}^{(i,l)} | \mathbf{x}^{(i,j)}) \| p(\mathbf{z}^{(i,l)}) \right) \bigg].
	\end{aligned}
\end{equation}
The first term in Eq. (\ref{eq:elbo-mivae}) is the sum of the reconstruction loss over all $d_j$ dimensions of sub-data $\mathbf{x}^{(i,j)}$, whilst the second term represents the KL-divergence between the posterior distribution $q(\mathbf{z}^{(i,j)} | \mathbf{x}^{(i,j)})$ and the prior distribution $p(\mathbf{z}^{(i,j)})$.

In practice, MIVAE maps the sub-data $\mathbf{}{x}^{(i,j)}$ into the latent representation $\mathbf{e}^{(i,j)}$ by using the $j^{th}$ sub-encoder, i.e., $\mathbf{e}^{(i,j)} = f(\mathbf{x}^{(i,j)}, \phi_{j})$, where $\phi_j = \{ \mathbf{W}^{(e,j)}, \mathbf{b}^{(e,j)} \}$ are weights and biases of the $j^{\emph{th}}$ sub-encoder. Next, $\overline{\mathbf{e}}^{(i)} = \mathbf{e}^{(i,1)} \bigoplus  \ldots \bigoplus \mathbf{e}^{(i,M)}$ is a combination of sub-data before $\boldsymbol{\mu}^{(i)} = f(\overline{\mathbf{e}}^{(i)}, \phi_{\mu})$ and $\boldsymbol{\sigma}^{(i)} = f(\overline{\mathbf{e}}^{(i)}, \phi_{\sigma})$ are calculated. $\phi_{\mu}$ and $\phi_{\sigma}$ are weights and biases of the neural networks, respectively. Like VAEAD, $\mathbf{z}^{(i, l)} = \boldsymbol{\mu}^{(i)} + \boldsymbol{\sigma}^{(i)}  \boldsymbol{\epsilon}^{(i,l)}$, where $\boldsymbol{\epsilon}^{(i,l)} \sim \mathcal{N}(0, I)$ and $l \in \{1, 2, \ldots, L\}$. Finally, the decoder of MIVAE aims to reconstruct the input $\mathbf{x}^{(i)}$ at its output, i.e., $\mathbf{\hat{x}}^{i} = f(\mathbf{z}^{(i)}, \theta)$, where $\theta = (\mathbf{W}^{(d)}, \mathbf{b}^{(d)})$ are weights and biases of the decoder. Note that, $\hat{\mathbf{x}}^{i} = \hat{\mathbf{x}}^{(i,1)}  \bigoplus \ldots \bigoplus \hat{\mathbf{x}}^{(i, M)}$, and the dimensionality of $\hat{\mathbf{x}}^{(i,j)}$ and $\mathbf{x}^{(i,j)}$ is equal. The MIVAE aims to maximize the
ELBO of a data sample $\mathbf{x}^{(i)}$, which is equivalently expressed as minimizing the loss function $-\mathcal{L}_{\emph{MIVAE}}^{(i)}$ in Eq. (\ref{eq:elbo-mivae}) as follows:
\begin{equation}
	\label{eq:loss_mivae}
	\begin{aligned}
		l_{\emph{MIVAE}}(\textbf{x}^{(i)}) = \frac{1}{L} \sum_{l=1}^{L} \sum_{j=1}^{M} \| \mathbf{x}^{(i,j)} - \hat{\mathbf{x}}^{(i,j,l)} \|^2   \\ + \frac{1}{2} \sum_{k=1}^{d_{z}} \big( 1 + \log ((\sigma_k^{(i)})^2) - (\mu_k^{(i)})^2 - (\sigma_k^{(i)})^2 \big) ,
	\end{aligned}
\end{equation} where $d_{z}$ is the dimensionality of $\mathbf{z}$. After training the MIVAE model, we define the anomaly function of sub-data sample $\mathbf{x}^{(i,j)}$ and data sample $\mathbf{x}^{(i)}$ as follows:
\begin{equation}
	\label{eq:score-MIVAE-xij}
	\begin{aligned}
		s_{\emph{MIVAE}}(\textbf{x}^{(i,j)}) = \frac{1}{L  d_j} \sum_{l=1}^{L}  \sum_{t=1}^{d_j} \| {x}^{(i,j)}_{t} - \hat{{x}}^{(i,j,l)}_{t} \|^2  ,
	\end{aligned}
\end{equation}
\begin{equation}
	\label{eq:score-MIVAE-xi}
	\begin{aligned}
		s(\textbf{x}^{(i)}) = \frac{1}{L} \sum_{l=1}^{L} \sum_{j=1}^{M} \left( \frac{1}{d_j}  \sum_{t=1}^{d_j} \beta^{(j)}  \| {x}^{(i,j)}_{t} - \hat{{x}}^{(i,j,l)}_{t} \|^2 \right) ,
	\end{aligned}
\end{equation} where $\beta^{(j)}$ is a balanced hyper-parameter which is calculated as follows:
\begin{equation}
	\label{eq:beta-overline}
	\begin{aligned}
		\overline{\beta} = \frac{1}{ M  N} \sum_{j=1}^{M} \sum_{i=1}^{N}  \left( s(\mathbf{x}^{(i,j)}) \right)  ,
	\end{aligned}
\end{equation}
\begin{equation}
	\label{eq:beta-j}
	\begin{aligned}
		\beta^{(j)} = \frac{1}{\overline{\beta}} \frac{1}{ N } \sum_{i=1}^{N} \left( s(\mathbf{x}^{(i,j)}) \right)    .
	\end{aligned}
\end{equation}
The anomaly score in Eq. (\ref{eq:score-MIVAE-xij}) aims to identify anomalies by using a feature subset of sub-data $\mathbf{x}^{(i,j)}$ of data sample $\mathbf{x}^{(i)}$, whilst Eq. (\ref{eq:score-MIVAE-xi}) is used to identify anomalies in the case of using all features of $\mathbf{x}^{(i)}$. Note that, for the training phase, we set $\beta^{(j)} = 1$ for $j=\{1, \ldots, M\}$. For the identifying anomaly phase after the training process, Eq. (\ref{eq:beta-j}) is used to calculate the anomaly score in Eq. (\ref{eq:score-MIVAE-xi}). $\overline{\beta}$ presents the average anomaly score across the entire feature space. $\beta^{(j)}$ presents the average anomaly score for a feature subset of the sub-dataset $\mathbf{X}^{(j)}$. If $M=1$ and $\beta^{(j)}=1$, the anomaly score of MIVAE in Eq. (\ref{eq:score-MIVAE-xi}) is the same as that of VAEAD in Eq. (\ref{eq:score_vae}).

%%%%%%%%%%%%%%%%%%%%%%%%%%%%%%%%%%%%%%%%%%%%%%%%%%%%%%
\section{Performance Evaluation}

%%%%%%%%%%%%%%%%%%%%%%%%%%%%%%%%%%%%%%%%%%%%%%%%%%%%%%
\subsection{Anomaly Detection Evaluation}
\label{l:MIVAE-proven}
This subsection discusses the performance of anomaly detection by using two anomaly score functions of VAEAD in Eq. (\ref{eq:score_vae}) and MIVAE in Eq. (\ref{eq:score-MIVAE-xi}) when they both use all features of $\mathbf{x}^{(i)}$. This is done by accepting the hypothesis as follows: 

\begin{hypothesis}
	\label{h:hypo_anomaly_score}
	In anomaly detection using reconstruction error from Auto-Encoder (AE) variants, anomalies exhibit higher reconstruction errors compared to normal samples, i.e., $ s_{\emph{AEAD}}(\mathbf{x}^{(i)}) < s_{\emph{AEAD}}(\mathbf{x}^{(k)}) $, where $ i = \{1, \ldots, n_b\} $ and $ k = \{n_b + 1, \ldots, n_b + n_a\} $ are the indices of the normal samples and anomalies in the dataset $ \mathbf{X}$, which can be separated by sub-datasets, i.e., $\mathbf{X} = \{\mathbf{X}^{(1)}, \mathbf{X}^{(2)}, \ldots, \mathbf{X}^{(M)} \}$.
\end{hypothesis}
The Hypothesis \ref{h:hypo_anomaly_score} was first introduced in \cite{sakurada2014anomaly}, and widely accepted via the experiments for specific datasets. We define the difference between the average score of the normal samples and anomalies as
\begin{equation}
	\label{eq:delta}
	\begin{aligned}
		\Delta{s_{(.)}} = s_{(.)}(\mathbf{x}^{(k)}) - s_{(.)}(\mathbf{x}^{(i)}), 
	\end{aligned}
\end{equation} where $(.)$ denotes the method used, e.g., AEAD, VAEAD, MIAEAD, and MIVAE. A large $\Delta{s_{(.)}}$ indicates that the anomaly score can effectively distinguish between anomalies from the normal samples, likely resulting in a higher AUC for anomaly identification.

\begin{theorem}
	\label{theo:theo_sub_score_ab}
	Assuming that $M$ sub-datasets of $\mathbf{X}$ have $M^{'}$ sub-datasets existing anomalies $j=\{1, \ldots, M^{'} \}$, where $ 1 \leq M^{'} \leq M$. The average anomaly score for each feature of normal samples in the dataset $\mathbf{X}$ is $\delta$, whilst that of anomalies is $E$. We will show that the difference in the average anomaly score between anomalies and normal samples for MIVAE is greater than that for VAEAD, i.e.,
	$\Delta{s_{\emph{MIVAE}}} \geq \Delta{s_{\emph{VAEAD}}}$.
\end{theorem}

\begin{proof}
    If $M=1$ and $\beta^{(j)}=1$, the anomaly score of MIVAE in Eq. (\ref{eq:score-MIVAE-xi}) is the same as that of VAEAD in Eq. (\ref{eq:score_vae}). To prove Theorem \ref{theo:theo_sub_score_ab}, we aim to show the difference in the average anomaly score between anomalies and normal samples for MIVAE when $M>1$ is greater than that for $M=1$ and $\beta^{(j)}=1$ (which is the same as VAEAD). To do this, we assume that the average anomaly score for each feature of normal samples in the dataset $\mathbf{X}$ for both MIVAE and VAEAD is $\delta$, whilst that of anomalies is $E$.
    
    Based on Eqs. (\ref{eq:score_vae}) and (\ref{eq:score-MIVAE-xi}), we aim to calculate: 
\begin{equation}
	\label{eq:delta-VAEAD}
	\begin{aligned}
        \Delta{s_{\emph{VAEAD}}} = \sum_{j=1}^{M^{'}} \big[ \frac{d_j}{d}  (E - \delta) \big], 
	\end{aligned}
\end{equation}
    \begin{equation}
	\label{eq:delta-MIVAE}
	\begin{aligned}
    \Delta{s_{\emph{MIVAE}}} = \sum_{j=1}^{M^{'}} \big[ \beta^{(j)}  (E - \delta)\big].
	\end{aligned}
\end{equation} To prove $\Delta{s_{\emph{MIVAE}}} \geq \Delta{s_{\emph{VAEAD}}}$ in Theorem \ref{theo:theo_sub_score_ab}, it is equivalent to proving 
\begin{equation}
	\label{eq:MIVAE>VAEAD}
	\begin{aligned}
    \sum_{j=1}^{M^{'}} \beta^{(j)} \geq \sum_{j=1}^{M^{'}} \frac{d_j}{d},
	\end{aligned}
\end{equation} where $E - \delta > 0$. The specific proof of Theorem \ref{theo:theo_sub_score_ab} is provided in Appendix \ref{l:proof-theorem1}.
\end{proof}
As observed in Theorem \ref{theo:theo_sub_score_ab}, the difference in the average anomaly score between normal samples and anomalies obtained by the proposed MIVAE is greater than that of the VAEAD, resulting in a higher AUC for MIVAE. Therefore, we may divide the datasets into sub-datasets to use as multiple inputs before applying MIVAE to identify anomalies.

%%%%%%%%%%%%%%%%%%%%%%%%%%%%%%%%%%%%%%%%%%%%%%%%%%%%%%

\subsection{Model Complexity}
\label{l_model_complexity}
This subsection discusses the parameters used by the MIVAE model compared to the fundamental model, i.e., VAEAD. Let $\{\alpha_1  d, \alpha_2  d, \ldots, \alpha_{T}  d \}$ be the number of neurons of the hidden layers of the encoder of AE, $T$ is the number of hidden layers of the encoder, and $  0 < \alpha_1, \ldots, \alpha_{T} < 1$.

\begin{lemma}
	\label{lem:lemma_parameter_encoder}
	Assuming that MIVAE and VAEAD have the same numbers of neurons at the $t^{th}$ and $({t +1})^{th}$ layers, the number of parameters used for fully connecting between the $t^{th}$ and $({t +1})^{th}$ layers of the encoder of the MIVAE model is lower than that of VAEAD. 
\end{lemma}

\begin{proof}
    The number of parameters used for fully connecting between the $t^{th}$ and $({t +1})^{th}$ layers of the encoder of the VAEAD and MIVAE models are: 
  \begin{equation}
	\begin{aligned}
        p_t(\emph{VAEAD})= (\alpha_t  d + 1)  \alpha_{t+1}  d = \alpha_t  \alpha_{t+1}  d^2 + \alpha_{t+1} d,
	\end{aligned}
\end{equation}
\begin{equation}
	\begin{aligned}
        p_t(\emph{MIVAE})= \sum_{j=1}^{M} \big((\alpha_t  d_j + 1)  \alpha_{t+1}  d_j \big), 
	\end{aligned}
\end{equation}
where $d_j$ is the number of neurons of the input layer of the $j^{th}$ sub-encoder of the encoder, and $M$ is the number of sub-encoders of MIVAE. 
We have shown that
\begin{equation}
    \label{eq:p_MIVAE>_VAEAD}
	\begin{aligned}
        p_t(\emph{VAEAD}) >  p_t(\emph{MIVAE}). 
	\end{aligned}
\end{equation}
The proof of Eq. (\ref{eq:p_MIVAE>_VAEAD}) is provided in Appendix \ref{l:lamda1}.
\end{proof}

\begin{lemma}
	\label{lem:lemma_parameter_decoder}
	The number of parameters used for fully connecting between the $t^{th}$ and $({t +1})^{th}$ layers of the decoder of MIVAE is the same as that of VAEAD. 
\end{lemma}
\begin{proof}
	Because the decoders of VAEAD and MIVAE have the same network architecture, the number of parameters used for fully connecting between the $t^{th}$ and $({t +1})^{th}$ layers of the decoders of VAEAD and MIVAE is equal to $\alpha_t  \alpha_{t+1}  d^2 + \alpha_{t} d$.
\end{proof}
From Lemma \ref{lem:lemma_parameter_encoder}, the number of parameters used for the encoder of MIVAE is lower than that of VAEAD. From Lemma \ref{lem:lemma_parameter_decoder}, VAEAD and MIVAE have the same parameters used in the decoder. Therefore, the number of parameters used for the MIVAE model is lower than that of the VAEAD model. Similarly, the number of parameters used for the MIAEAD model is lower than that of the Auto-Encoder model.

%%%%%%%%%%%%%%%%%%%%%%%%%%%%%%%%%%%%%%%%%%%%%%%%%%%%%%
\subsection{Training Evaluation}
\label{l_training_evaluation}

%%%%%%%%%%%%%%%%%%%%%%%%%%%%%%%%%%%%%%%%%%%%%%%%%%%%%%
\subsubsection{Convergence Discussion}
\begin{table}[t] %\vspace{0.02}
	\caption{Comparison of the feedforward pass and backpropagation of the encoder in VAEAD and MIVAE.}
	\label{tab:forward-backward}
	\setlength\tabcolsep{5pt}
	\centering
	%\small
	\scriptsize
	%\footnotesize	
    \begin{tabular}{|l|l|l|}
    \hline
    \multirow{2}{*}{Forward}  & VAEAD   & \scriptsize $h_{i}^{(t)} = g\left(\sum_{k=1}^{d^{(t-1)}} {W_{ik}^{(t-1)}} h_{k}^{(t-1)} + b_{i}^{(t-1)}\right)$ \\ \cline{2-3} 
    & MIVAE & $h_{ij}^{(t)} = g\left(\sum_{k=1}^{d^{(t-1)}_{j}} {W_{ijk}^{(t-1)}} h_{jk}^{(t-1)} + b_{ij}^{(t-1)}\right)$ \\ \hline
    \multirow{2}{*}{Backward} & VAEAD   & $\sigma_{k}^{(t)} = \sum_{i=1}^{d^{(t+1)}} W_{ik}^{(t)} \sigma_{i}^{(t+1)} \odot g'\left(h_{k}^{(t)}\right)
$
 \\ \cline{2-3} 
    & MIVAE & $\sigma_{jk}^{(t)} = \sum_{i=1}^{d^{(t+1)}_{j}} W_{ijk}^{(t)} \sigma_{ij}^{(t+1)} \odot g'\left(h_{jk}^{(t)}\right)
$ \\ \hline
    \end{tabular}
    
\end{table}
To discuss the convergence of MIVAE, we first show that MIVAE is an advanced version of VAEAD \cite{an2015variational}. The key difference is that MIVAE processes feature subsets using sub-encoders and aims to reconstruct these separated feature subsets simultaneously, rather than reconstructing all features together of the VAEAD. As observed in Table \ref{tab:forward-backward}, $h_{i}^{(t)}$ is the activation of node $i$ in the layer $t$ of the encoder of VAEAD, whilst that of MIVAE for node $i$ in the sub-encoder $j$ is $h_{ij}^{(t)}$. 
$\sigma_{k}^{(t)}$ is error term for node $k$ in layer $t$, and $g'(h_{k}^{(t)})$ is the derivative of the activation function applied to $h_{i}^{(t)}$. $\odot$ is an element-wise operation. $d^{(t)}$ is the number of neurons in layer $t$ of VAEAD, and $d^{(t)}_j$ is the number of neurons in layer $t$ and sub-encoder $j$ of MIVAE.
The gradients of the weights and biases of VAEAD and MIVAE are $\nabla W_{ik}^{(t)} = \sigma_i^{(t+1)} h_k^{(t)}, \nabla b_i^{(t)} = \sigma_i^{(t+1)}
$  and $\nabla W_{ijk}^{(t)} = \sigma_{ij}^{(t+1)} h_{jk}^{(t)}, \nabla b_{ij}^{(t)} = \sigma_{ij}^{(t+1)}$, respectively. Therefore, the feedforward pass and backpropagation of the VAEAD and MIVAE encoders are calculated similarly. In addition, the loss functions of VAEAD in Eq. (\ref{eq:score_vae}) and MIVAE in Eq. (\ref{eq:score-MIVAE-xi}) are both mean squared error functions. As a result, the training processes of MIVAE and VAEAD are identical.
Next, we discuss the convergence of MIVAE when applying the ADAM optimization \cite{adam}. Based on the result in \cite{rmsprop}, the MIVAE optimized using ADAM will converge to the critical points under the assumption that the loss function in Eq. (\ref{eq:loss_mivae}), i.e., $l_{\emph{MIVAE}}^{(i)}(\phi_1,\ldots,\phi_M, \theta)$, and its gradient are Lipschitz continuous and lower-bounded. Here, $K$ refers to the total number of iterations. In addition, if the learning rate decreases as $\eta_k = \frac {\eta_1}{\sqrt{k}}$ and $\beta_1 < \beta_2 < 1$, the gradient of the loss function of MIVAE after $K$ iterations is bounded as $\min_{t \in (1, K]} \| \nabla f_k(\phi_1,\ldots,\phi_M, \theta) \|_1 \leq O \left( \frac{\log K}{\sqrt{K}} \right)$. This implies that the optimizer's gradient will eventually approach zero at a rate of $O \left( \frac{\log K}{\sqrt{K}} \right)$. Therefore, when Adam is applied to optimize MIVAE, the algorithm will converge to stationary points.

%%%%%%%%%%%%%%%%%%%%%%%%%%%%%%%%%%%%%%%%%%%%%%%%
\begin{table}[t] %\vspace{0.02}
	\caption{Comparison complexity of training process of VAEAD and MIVAE.}
	\label{tab:training-complexity}
	\setlength\tabcolsep{0pt}
	\centering
	%\small
	\scriptsize
	%\footnotesize	
	\begin{tabular}{|l|ll|}
		\hline
		\multicolumn{1}{|c|}{Operation} & \multicolumn{1}{c|}{VAEAD} & \multicolumn{1}{c|}{MIVAE} \\  \hline
		Encoder          & \multicolumn{1}{l|}{ \scriptsize $O \left(  N  d^2  \sum_{t=1}^{T-1} \left( \alpha_t  \alpha_{t+1} \right) \right)
			$}      &\scriptsize $O\left( N  \left(\sum_{j=1}^{M} \left( d_j^2  \sum_{t=1}^{T-1} \alpha_t \alpha_{t+1}\right)  \right) \right)
		$
		\\ \hline
		Sampling & \multicolumn{2}{c|}{\scriptsize $O\left( L N  d_z \right )$}              \\ \hline
		Decoder          & \multicolumn{2}{c|}{\scriptsize $O \Big(  L  N  d^2 \sum_{t=1}^{T-1} (\alpha_t  \alpha_{t+1}) \Big)  $ }  
		\\ \hline
		Loss    & \multicolumn{1}{l|}{\scriptsize $ O(N  d + N  d_z)$
		}      & \scriptsize $O\left( L  N  \sum_{j=1}^{M} d_j +  L  N  d_z \right)$
		\\ \hline
	\end{tabular}
	
\end{table}

\subsubsection{Training Complexity}
We evaluate the training process for one epoch. The training complexity of the fully connected neural network depends on four steps: forward propagation, backpropagation, gradient calculation for the loss function, and loss function computation. Each neuron computes a weighted sum of its inputs in forward propagation and then applies an activation function. In backpropagation, the error is propagated backwards from the output to the previous layers to update the weights and biases of each neuron. Therefore, forward propagation, back-propagation, and gradient calculation complexity depend on the number of neurons in each layer and the connections between neurons in two consecutive layers. The difference in training complexity for VAEAD and MIVAE arises from their encoders and loss functions, while the sampling process and complexity of training the decoders are the same for both models. 

We first consider the complexity of training one data sample. For VAEAD, the complexity of forward propagation and back-propagation between the $t^{th}$ and $(t+1)^{th}$ layers is $O(\alpha_t d  \alpha_{t+1}  d) = O(d^2  \alpha_t  \alpha_{t+1})$. For $T$ layers of the encoder, the complexity is $O\left( d^2  \sum_{t=1}^{T-1} (\alpha_t  \alpha_{t+1})\right)$. For MIVAE, the complexity of the $j^{th}$ sub-encoder for training process is $O\left( d{_j}^2  \sum_{t=1}^{T-1} (\alpha_t  \alpha_{t+1})\right)$. For $M$ sub-encoders, the complexity is $O\left( \sum_{j=1}^{M} \left( d{_j}^2  \sum_{t=1}^{T-1} (\alpha_t  \alpha_{t+1})\right) \right)$. For the sampling process of both VAEAD and MIVAE, the complexity is $O(L d_z)$, where $L$ is the number of sampling $z$. For their decoder, the complexity is $O\left(L  d^2  \sum_{t=1}^{T-1} (\alpha_t  \alpha_{t+1})\right)$. For loss function calculation, the complexity for VAEAD and MIVAE is $O(d + d_z)$ and $O(L \sum_{j=1}^{M} d_j + L  d_z)$, respectively. 

Finally, we expect the complexity of the training process of VAEAD and MIVAE for $N$ samples of dataset $\mathbf{X}$, as shown in Table \ref{tab:training-complexity}. The complexity of training VAEAD and MIVAE is $o(L  N d^2  T)$ and $o(L  N d^2  T + M N d^2  T)$, respectively. When $M=d$, it implies that each sub-encoder can input a feature, resulting in $o(L  N d^2  T + N d^3  T)$ of the training complexity of MIVAE.
Therefore, the complexity of the MIVAE training process is higher than that of VAEAD.
As the number of sub-encoders in MIVAE increases, the training time of MIVAE will be much longer than that of VAEAD. Similarly, it is observed that the training time of MIAEAD will be longer than that of AEAD by removing the complexity of the sampling process of VAEAD and MIVAE.

%%%%%%%%%%%%%%%%%%%%%%%%%%%%%%%%%%%%%%%%%%%%%%%%%%%%%%

%%%%%%%%%%%%%%%%%%%%%%%%%%%%%%%%%%%%%%%%%%%%%%%%%%%%%%

\section{Experimental Setups}
\subsection{Performance Metrics}
To evaluate the performance of the anomaly detection models, we define $\emph{TP}$, $\emph{TN}$, $\emph{FP}$, and $\emph{FN}$ as \textit{True Positive}, \textit{True Negative}, \textit{False Positive}, and \textit{False Negative}, respectively. The Receiver Operating Characteristic (ROC) curve is a plot of the $\emph{TP}$ against the $\emph{FP}$ for different threshold values. Similar to \cite{AEWAD}, 
we use the area under the ROC curve ($\emph{AUC}$) to evaluate the performance of anomaly detection models. To calculate the $\emph{AUC}$, we use anomaly scores, i.e., $s^{(i)}$, and the ground truth labels. This is because the anomaly scores can be used to create a binary classification that $s^{(i)}$ greater than a threshold is predicted as an anomaly, while others are labelled as normal. Note that, to calculate the $\emph{AUC}$, we do not need to know the number of anomalies in the dataset. To further evaluate the performance of the anomaly detection models, we use three scores, i.e., $\emph{F-score}$, Miss Detection Rate ($\emph{MDR}$) and  False Alarm Rate ($\emph{FAR}$), where $\emph{MDR} =\frac{\emph{FN}}{\emph{FN} + \emph{TP}}$  and $\emph{FAR} = \frac{\emph{FP}}{\emph{FP}+\emph{TN}}$. To calculate $\emph{F-score}$, $\emph{FAR}$ and $\emph{MDR}$, we assign $n_a$ samples that have the largest anomaly scores as anomalies while others are predicted as normal labels. In addition, we measure the heterogeneity of each dataset by using the coefficient of variation ($\emph{CV}$) \cite{verrill2007confidence}, where $\emph{CV} = 100  \frac{1}{d} \sum_{t=1}^{d} \frac{\sigma_{t}}{\mu_{t}} $, $\mu_{t} > 0$. Here, $\sigma_{t}$, $\mu_{t}$, and $d$ represent the standard deviation, the mean of the $t^{th}$ feature, the dimensionality of dataset, respectively. A high $\emph{CV}$ indicates large variability compared to the mean, whilst a low $\emph{CV}$ indicates small variability around the mean. In other words, a high $\emph{CV}$ suggests that the dataset is highly heterogeneous.

\subsection{Datasets}
We evaluate the performance of anomaly detection models on eight datasets, i.e., cardio disease (M1), credit card fraud (M2) from Kaggle, Arrhythmia (M3),  medical image Mammography (M4), network security NSLKDD (M5),  satellite image (M6), Shuttle (M7), and Spambase (M8) and from the UCI repository \cite{AEWAD}. The numbers of samples $N$, dimensionality $d$, numbers of anomalies $n_a$, numbers of benign samples $n_b$, and the ratio of the number of anomalies over the number of all samples $\frac{n_a}{N}$, of these datasets are shown in Table \ref{tab:datasets}. The heterogeneity of each dataset is calculated by the coefficient of variation ($\emph{CV}$).

\subsection{Experimental Setting}

\begin{table}[t] %\vspace{0.02}
	\caption{Datasets.}
	\label{tab:datasets}
	\setlength\tabcolsep{1.5pt}
	\centering
	%\small
	\scriptsize
	%\footnotesize	
	\begin{tabular}{|c|c|c|c|c|c|c|c|}
		\hline
        \multicolumn{2}{|c|}{Datasets' Name} & $N$      & $n_a$  & $n_b$   & $\frac{n_a}{N}$  & $d$   & $CV$   \\ \hline
		Cardio      & M1   & 1830   & 176   & 1654   & 9.6\%  & 21  & 25.6 \\ \hline
		Fraud       & M2   & 284806 & 492   & 284314 & 0.2\%  & 29  & 8.8  \\ \hline
		Arrhythmia  & M3   & 451    & 65    & 386    & 14.4\% & 279 & 38.4 \\ \hline
		Mammography & M4   & 11182  & 260   & 10922  & 2.3\%  & 6   & 42.9 \\ \hline
		NSL         & M5   & 148516 & 77053 & 71463  & 51.9\% & 122 & 87.4 \\ \hline
		Satellite   & M6   & 6434   & 2036  & 4398   & 31.6\% & 36  & 8.0  \\ \hline
		Shuttle     & M7   & 49096  & 3510  & 45586  & 7.1\%  & 9   & 4.3  \\ \hline
		Spambase    & M8   & 4600   & 1812  & 2788   & 39.4\% & 57  & 22.1 \\ \hline
	\end{tabular}
\end{table}

\begin{table}[t]
	\caption{Hyper-parameters tuning.}
	\label{tab:hyper-parameters}
	\setlength\tabcolsep{1.5pt}
	\centering
	%\small
	\scriptsize
	%\footnotesize	
	\begin{tabular}{|l|l|}
		\hline
		LoF     & $n\_\emph{neighbor}$ =\{2,4,6,8,10,15,20,50\} \\ \hline
		KDE     & $\emph{bandwidth}$=\{0.001, 0.005, 0.1, 0.2, 0.4, 0.5, 1.0, 2.0, 5.0, 10, 20\}    \\ \hline
		OSVM    & $\gamma$=\{0.001, 0.01, 0.05, 0.08, 0.1, 0.2, 0.5, 1\}\\ \hline
		RDAE    & $\lambda$ =\{0.001, 0.005, 0.01, 0.05, 0.1, 0.5, 0.8, 1, 2, 5, 8, 10, 20\}   \\ \hline
		RandNet & $n\_\emph{model}$=\{2,3,5,7,9,11,15,20,30,35,40,45,50\}   \\ \hline
		AE      & $d_z$ = \{0.2, 0.4, 0.6, 0.8\}        \\ \hline
		VAE     & \begin{tabular}[c]{@{}l@{}}  $L$=\{1,5,10,20,50\};\\ $d_z$=\{0.2, 0.4, 0.6, 0.8, 1, 1.2, 1.5, 2, 5\}\end{tabular} \\ \hline
		AnoGAN  & $\alpha$ = \{0.1, 0.2, 0.3, 0.4, 0.5, 0.6, 0.7, 0.8, 0.9\}        \\ \hline
		SkipGAN &  $\alpha$=\{0.1, 0.2, 0.3, 0.4, 0.5, 0.6, 0.7, 0.8, 0.9\}        \\ \hline
		AAE     &  $\alpha$=\{0.1, 0.2, 0.3, 0.4, 0.5, 0.6, 0.7, 0.8, 0.9\}        \\ \hline
		MIAEAD  & $n\_{\emph{branch}}$=\{3,5,7,9,15,25,35\}       \\ \hline
		MIVAE   & \begin{tabular}[c]{@{}l@{}} $n\_{\emph{branch}}$=\{3,5,7,9,15,25,35\}\\ $L$=\{1,5,10,20,50\}\end{tabular}     \\ \hline
	\end{tabular}
	
\end{table}

We use grid search to tune the hyper-parameters of anomaly detection models, as observed in Table \ref{tab:hyper-parameters}. For LOF, KDE \cite{tang2017local}, and OSVM \cite{OSVM}, the hyper-parameters are $n\_neighbor$, $bandwidth$, and $\gamma$, respectively. $n\_neighbor$ is the number of nearest neighbours used to compute the local reachability density. We use the Gaussian kernel to set the KDE method, and the $bandwidth$ is the bandwidth of the kernel. $\gamma$ is an important hyper-parameter used for OSVM to determine the spread of the Gaussian kernel. Next, RDA uses $\lambda$ to tune the level of sparsity of noise and outliers in the original dataset \cite{RDAE}. A small value of $\lambda$ encourages much of the data to be isolated in the set of noise/outliers to minimize the reconstruction error while the large $\lambda$ increases the reconstruction error of the AE. 
For RandNet, the number of models randomly generated to form an ensemble of autoencoders is denoted as $n_{model}$ \cite{RandNet}. For DAGMM, we refer to the results from \cite{AEWAD}. For AEAD in \cite{pang2021deep}, the dimensionality of the latent space, $d_z$, is tuned. Similarly, for VAEAD in \cite{an2015variational}, we tune both the number of samples drawn $L$ and the latent dimensionality $d_z$. In the case of AnoGAN \cite{AnoGAN}, SkipGAN \cite{SkipAnoGAN}, and AAE \cite{AAE}, the hyperparameter $\alpha$ is used to balance the generator's reconstruction error and the reconstruction error in the discriminator's latent space.

To conduct experiments on neural networks, i.e., AEAD, VAEAD, RDAE, RandNet, AnoGAN, SkipGAN, AAE, MIAEAD, and MIVAE, we use Tensorflow and the Adam optimization algorithm to train the model \cite{dinh2024multipleinput}. The numbers of batch size, epoch, and learning rate are 100, 5000, and $10^{-4}$, respectively. The number of sub-encoders used is $n\_sub\_encoder$, as observed in Table \ref{tab:hyper-parameters}. Similar to authors of \cite{dinh2024multipleinput}, the numbers of neurons of layers are the list $\{ d, 0.8  d, 0.6  d, 0.4  d, d_z, 0.4  d,  0.6  d,  0.8  d, d \}$, where $d$ is the dimensionality of the input data, $d_z = \lfloor \sqrt{d} \rfloor + 1$ is the number of neurons of bottleneck layer of the MIAEAD. For MIVAE, the number of neurons of layers is the list $\{ d, 0.9  d, 0.8  d, 2  d_z,  0.8  d,  0.9  d, d \}$. We consider that there are $M$ sub-datasets in the dataset, and the dimensionality of each sub-dataset is nearly the same. $M$ is tested by the list $\{3, 5, 7, 9, 15, 25, 35\}$, as presented in Table \ref{tab:hyper-parameters}. After setting the number of neurons of the encoder, we can calculate the number of neurons of each hidden layer of each sub-encoder. For example, if the dimensionality of the input of the M2 dataset is $d=29$, we can split the list $\{ 10,10,9\}$ into three sub-encoders of the MIAEAD and MIVAE. In addition, we use the Tanh active function for every layer to configure the MIAEAD and MIVAE, while the last layers of the Encoder and Decoder use the Relu active function. Note that the $\emph{AUC}$ of MIAEAD and MIVAE is the maximum value across all sub-encoders. To obtain the best $\emph{AUC}$ for AnoGAN, SkipGAN, and AAE, we select the maximum value among their three anomaly scores: the reconstruction error of the generator, the reconstruction error in the latent space of the discriminator, and the sum of both reconstruction errors.

\section{Experiment Results}
\label{l_experiment_results}
\subsection{Performance of Anomaly Detection Models}

We present the main results of MIAEAD and MIVAE compared to other methods.
First, we compare the $\emph{AUC}$ of MIAEAD and MIVAE with anomaly detection methods, i.e., LOF, KDE \cite{tang2017local}, SVM \cite{OSVM}, AEAD \cite{pang2021deep}, RandNet \cite{RandNet}, RDA \cite{RDAE}, and DAGMM \cite{DAGMM}. As observed in Table \ref{tab:experimental_results_01}, the $\emph{AUC}$  obtained by MIAEAD and MIVAE is significantly greater than other methods over eight datasets. In addition, the average $\emph{AUC}$ obtained by MIAEAD and MIVAE is greater than those of the state-of-the-art anomaly detection methods, i.e., DAGMM and RandNet, by up to $4.3\%$ and $6\%$, respectively. For example, MIAEAD and MIVAE achieve average of $0.866$ and $0.883$ in terms of $\emph{AUC}$ on the M3 dataset compared to $0.0.626$, $0.464$, $0.576$, $0.732$, $0.738$, $0.614$, and $0.823$, for LOF, KDE, SVM, AE, RandNet, RDA, and DAFGMM, respectively. Three methods, e.g., LOF, KDE, and OSVM, report low $\emph{AUC}$ over eight datasets since these conventional methods are likely to suffer from high-dimensional problems of non-IID data and a high value of heterogeneity of datasets ($CV$). Similarly, the RDA significantly depends on the hyper-parameter selected, i.e., $\lambda$, causing low values of $\emph{AUC}$. AEAD and RandNet present low values of $\emph{AUC}$  on the M5 dataset because their ratio of $\frac{n_a}{N}$ is large.

Second, we compare the results of MIVAE to those of the generative models, including VAEAD, AAE, AnoGAN, and SkipGAN, as shown in Table \ref{tab:experimental_results_02}. MIVAE uses feature subsets to identify anomalies based on Eq. (\ref{eq:score-MIVAE-xij}). MIVAE achieves the $\emph{AUC}$ by selecting the maximum value from feature subsets corresponding to sub-encoders. In addition, AAE, AnoGAN, and SkipGAN obtain their $\emph{AUC}$ values by selecting the maximum value from three anomaly score functions, i.e., the generator's reconstruction error, the discriminator's reconstruction error in the latent space, and the sum of both.
In general, MIVAE achieves a greater $\emph{AUC}$ than the generative models. For example, the average $\emph{AUC}$ obtained by VAEAD, AAE, AnoGAN, SkipGAN, and MIVAE is $0.716$, $0.808$, $0.776$, $0.883$, and $0.898$, respectively. These results highlight the superior performance of MIVAE over the other methods, as MIVAE can identify anomalies using feature subsets instead of relying on the entire feature space.

%%%%%%%%%%%%%%%%%%%%%%%%%%%%%%%%%%%%%%%%%%%
\begin{table}[t]
	\caption{Performance of MIAEAD and MIVAE compared to the other conventional unsupervised methods.% We take the results of DAGMM from \cite{AEWAD}.
	}
	\label{tab:experimental_results_01}
	\setlength\tabcolsep{1.5pt}
	\centering
	%\small
	\scriptsize
	%\footnotesize	
	\begin{tabular}{|c|c|c|c|c|c|c|c|c|c|}
		\hline 
		Datasets & LOF   & KDE   & OSVM  & AEAD    & Rnet & RDAE  & DAGMM& MIAEAD         & MIVAE         \\ \hline
		M1       & 0.743 & 0.445 & 0.595 & 0.761 & \textbf{0.969} & 0.544 & 0.918& 0.921& 0.947\\ \hline
		M2       & 0.519 & 0.350 & 0.502 & 0.950 & 0.950& 0.874 & 0.942& 0.951& \textbf{0.953} \\ \hline
		M3       & 0.792 & 0.500 & 0.602 & 0.742 & 0.652& 0.533 & 0.736& 0.803& \textbf{0.817} \\ \hline
		M4       & 0.821 & 0.267 & 0.533 & 0.866 & 0.872& 0.772 & 0.838& 0.878& \textbf{0.888} \\ \hline
		M5       & 0.507 & 0.776 & 0.698 & 0.267 & 0.214& 0.500 & 0.813& 0.831& \textbf{0.898} \\ \hline
		M6       & 0.565 & 0.480 & 0.586 & 0.676 & 0.647& 0.522 & \textbf{0.770} & 0.702& 0.755\\ \hline
		M7       & 0.566 & 0.298 & 0.569 & 0.987 & 0.990& 0.587 & \textbf{0.992} & 0.987& 0.988\\ \hline
		M8       & 0.498 & 0.595 & 0.522 & 0.607 & 0.609& 0.582 & 0.577& \textbf{0.857} & 0.821\\ \hline \hline
		\rowcolor[HTML]{DDDDDD} 
		Average      & 0.626 & 0.464 & 0.576 & 0.732 & 0.738& 0.614 & 0.823& 0.866& \textbf{0.883} \\ \hline
	\end{tabular}
	
\end{table}

\begin{table}[t]
	\caption{Performance of MIVAE compared to generative methods.
	}
	\label{tab:experimental_results_02}
	\setlength\tabcolsep{1.5pt}
	\centering
	%\small
	\scriptsize
	%\footnotesize	
	\begin{tabular}{|c|c|c|c|c|c|}
		\hline 
		Datasets & VAEAD   & AAE & AnoGAN & SkipGAN & MIVAE         \\ \hline \hline
		M1       & 0.935 & 0.848& 0.832   & 0.858    & \textbf{0.947} \\ \hline
		M2       & 0.949 & 0.952& 0.928   & 0.934    & \textbf{0.953} \\ \hline
		M3       & 0.802 & 0.795& 0.777   & 0.761    & \textbf{0.817}         \\ \hline
		M4       & 0.860 & 0.882& 0.786   & 0.821    & \textbf{0.888}\\ \hline
		M5       & 0.051 & 0.595& 0.632   & 0.769    & \textbf{0.898} \\ \hline
		M6       & 0.606 & 0.684& 0.652   & 0.737    & \textbf{0.755} \\ \hline
		M7       & 0.986 & \textbf{0.990} & 0.951   & 0.969    & 0.988\\ \hline
		M8       & 0.539 & 0.715& 0.654   & 0.658    & \textbf{0.821} \\ \hline \hline
		\rowcolor[HTML]{DDDDDD} 
		Average  & 0.716 & 0.808& 0.776   & 0.813    & \textbf{0.883} \\ \hline       
	\end{tabular}
	
\end{table}

\begin{table}[!t]
	\caption{FAR and MDR obtained by MIAEAD and MIVAE.}
	\label{tab:experimental_results_04}
	\setlength\tabcolsep{1.5pt}
	\centering
	%\small
	\scriptsize
	%\footnotesize	
	\begin{tabular}{|c|c|c|c|c|c|c|c|c|c|c|}
		\hline
		scores      & Methods & M1    & M2    & M3    & M4    & M5    & M6    & M7    & M8    & \textbf{Average} \\ \hline \hline
		\multirow{2}{*}{Fscore} & MIAEAD  & 0.924 & 0.999 & 0.854 & 0.973 & 0.769 & 0.682 & 0.933 & 0.807 & 0.868\\ \cline{2-11} 
		& MIVAE   & 0.952 & 0.999 & 0.832 & 0.973 & 0.819 & 0.717 & 0.993 & 0.764 & \textbf{0.881}   \\ \hline \hline
		\multirow{2}{*}{FAR}    & MIAEAD  & 0.364 & 0.408 & 0.439 & 0.534 & 0.228 & 0.417 & 0.073 & 0.212 & 0.334\\ \cline{2-11} 
		& MIVAE   & 0.229 & 0.390 & 0.506 & 0.560 & 0.179 & 0.371 & 0.048 & 0.258 & \textbf{0.318}   \\ \hline \hline
		\multirow{2}{*}{MDR}    & MIAEAD  & 0.076 & 0.001 & 0.146 & 0.025 & 0.231 & 0.318 & 0.011 & 0.193 & 0.125\\ \cline{2-11} 
		& MIVAE   & 0.048 & 0.001 & 0.168 & 0.027 & 0.181 & 0.283 & 0.007 & 0.236 & \textbf{0.119}   \\ \hline
	\end{tabular}
	
\end{table}

%%%%%%%%%%%%%%%%%%%%%%%%%%%%%%%%%%%%%%%%%%%

Third, we report the $\emph{Fscore}$, $\emph{FAR}$, and $\emph{MDR}$ obtained by MIAEAD and MIVAE in Table \ref{tab:experimental_results_04}. The results obtained by MIVAE are greater than those of MIAEAD across all three metrics. For example, MIAEAD achieves an average of $0.868$, $0.334$, and $0.125$ for $\emph{Fscore}$, $\emph{FAR}$, and $\emph{MDR}$ over eight datasets, respectively. For MIVAE, the results are $0.881$, $0.318$, and $0.119$ in terms of average $\emph{Fscore}$, $\emph{FAR}$, and $\emph{MDR}$ over eight datasets. This demonstrates the benefit of modelling the distribution of normal data in the latent space of MIVAE compared to MIAEAD which does not learn the distribution of data in the latent space.

Overall, the above results demonstrate the superior performance of MIAEAD and MIVAE over conventional anomaly detection methods, such as LOF, KDE, and OSVM. The $\emph{AUC}$ achieved by MIAEAD and MIVAE is significantly greater than that of AEAD and RDA. The $\emph{AUC}$ of MIAEAD and MIVAE outperforms state-of-the-art methods, such as RandNet and DAGMM, by up to $6\%$, respectively. MIVAE also achieves a higher $\emph{AUC}$ compared to generative models, such as VAEAD, AAE, AnoGAN, and SkipGAN, for anomaly detection. This is because MIAEAD and MIVAE can detect anomalies using feature subsets instead of relying on the entire feature space. The $\emph{AUC}$ of MIVAE is greater than that of MIAEAD due to its ability to model the distribution of normal data in the latent space, enabling it to identify anomalies that significantly deviate from the normal distribution. The superiority of MIAEAD over other methods is further explained in the next subsection.

%%%%%%%%%%%%%%%%%%%%%%%%%%%%%%%%%%%%%%%%%%%
\begin{table*}[!t]
	\centering
	\begin{center} \hspace*{-0.3cm} 
		\begin{tabular}{m{0em}cc} 
			&\begin{subfigure}{0.4\textwidth}\centering\includegraphics[width=\linewidth]{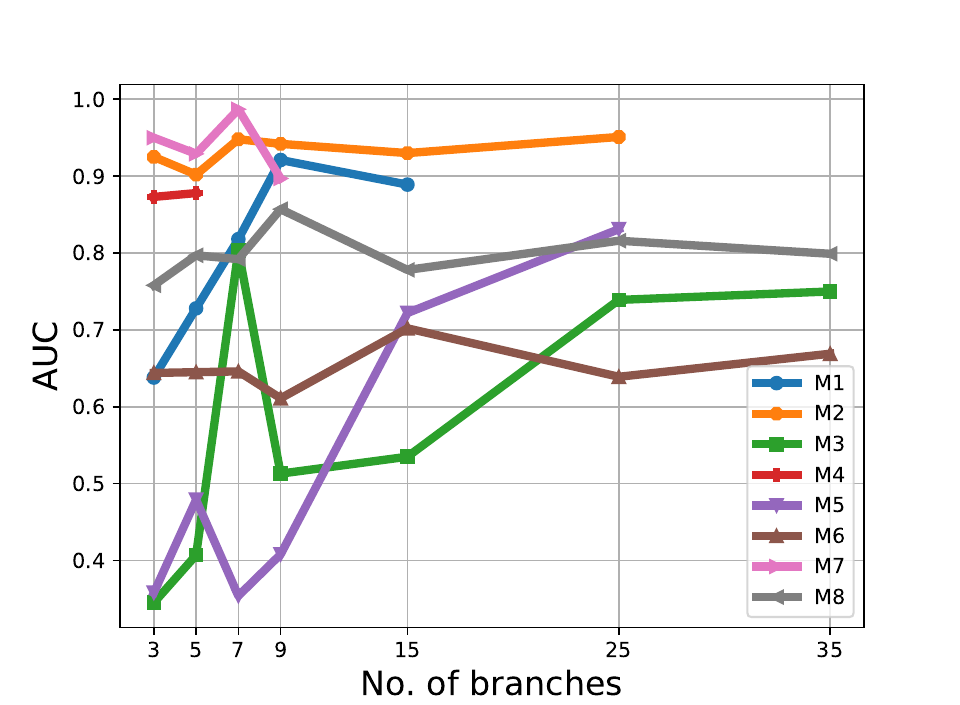}\end{subfigure}
			&\begin{subfigure}{0.4\textwidth}\centering\includegraphics[width=\linewidth]{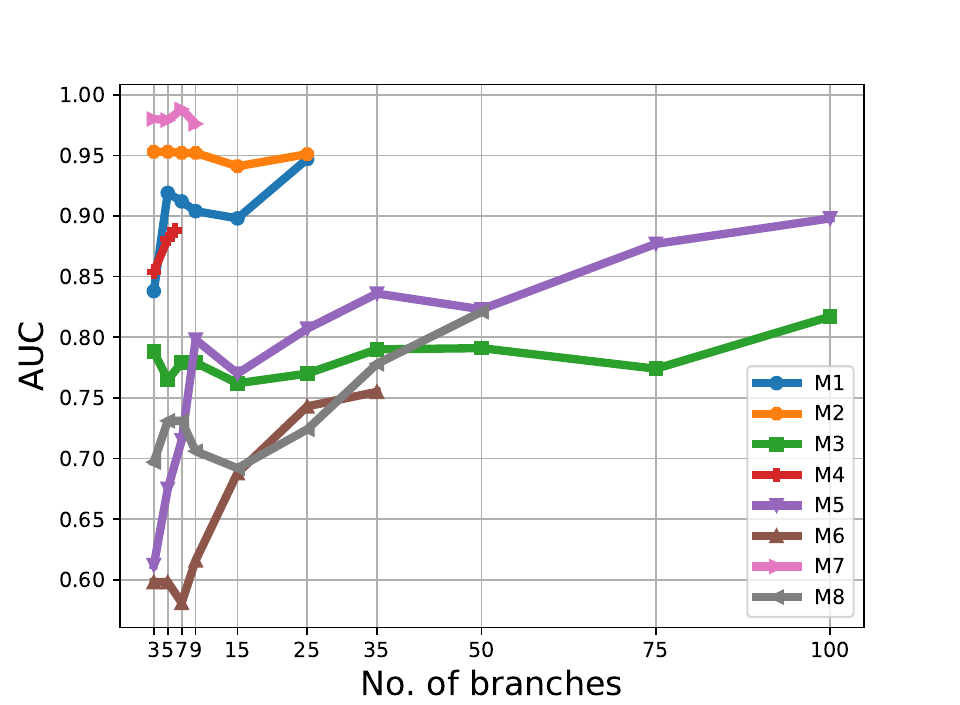}\end{subfigure}
			\\
			& (a) MIAEAD & (b) MIVAE \\
			\\
		\end{tabular}
	\end{center} \vspace*{-0.5 cm}
	\captionof{figure}{$\emph{AUC}$ obtained by MIAEAD and MIVAE on numbers of sub-encoders used.}
	\label{fig:results-No-branches-used}
\end{table*}

\begin{table}[t]
	\caption{Performance of MIVAE compared to generative methods when using a sum of anomaly score.
	}
	\label{tab:experimental_model_01}
	\setlength\tabcolsep{1.5pt}
	\centering
	%\small
	\scriptsize
	%\footnotesize	
	\begin{tabular}{|c|c|c|c|c|c|}
		\hline 
		Datasets & VAEAD   & AAE*& AnoGAN*         & SkipGAN* & MIVAE*\\ \hline \hline
		M1       & 0.935 & 0.848& 0.802& 0.858   & \textbf{0.960} \\ \hline
		M2       & 0.949 & 0.952& 0.916& 0.920   & \textbf{0.955} \\ \hline
		M3       & 0.802 & 0.795& 0.741& 0.742   & \textbf{0.808} \\ \hline
		M4       & 0.860 & 0.882& 0.786& 0.813   & \textbf{0.902} \\ \hline
		M5       & 0.051 & 0.374& \textbf{0.628} & 0.408   & 0.473\\ \hline
		M6       & 0.606 & 0.652& 0.625& 0.616   & \textbf{0.670} \\ \hline
		M7       & 0.986 & \textbf{0.990} & 0.933& 0.964   & \textbf{0.990} \\ \hline
		M8       & 0.539 & \textbf{0.715} & 0.530& 0.512   & 0.616\\ \hline \hline
		\rowcolor[HTML]{DDDDDD} 
		Average      & 0.716 & 0.776& 0.745& 0.729   & \textbf{0.797} \\ \hline
	\end{tabular}
	
\end{table}

%%%%%%%%%%%%%%%%%%%%%%%%%%%%%%%%%%%%%%%%%%%

\subsection{Model Analysis of MIAED and MIVAE}
\label{Analysis_MIAEAD}

First, we compare the $\emph{AUC}$ obtained by MIVAE with that of generative models, such as VAEAD, AAE, AnoGAN, and SkipGAN when MIVAE uses all features to identify anomaly, as shown in Table \ref{tab:experimental_model_01}. While MIVAE uses the anomaly function defined in Eq. (\ref{eq:score-MIVAE-xi}), AAE, AnoGAN, and SkipGAN use the sum of the anomaly scores based on the generator's reconstruction error and the discriminator's reconstruction error in the latent space \cite{AAE, AnoGAN, SkipAnoGAN}. This means that MIVAE focuses on using all feature space instead of feature subsets to identify anomalies. The $\emph{AUC}$ obtained by MIVAE is significantly greater than that of the generative models. For example, MIVAE, VAEAD, AAE, AnoGAN, and SkipGAN achieve $0.716$, $0.776$, $0.745$, $0.729$, and $0.797$, respectively, in terms of average $\emph{AUC}$ across eight datasets. This demonstrates that MIVAE effectively identifies anomalies by leveraging all feature space instead of only using feature subsets, as proven in subsection \ref{l:MIVAE-proven}.

Second, we evaluate the $\emph{AUC}$ obtained by MIAEAD and MIVAE in two methods. In the first method, called MIAEAD-max and MIVAE-max, we calculate the  $\emph{AUC}$ of each sub-encoder of the $j^{th}$ sub-dataset $\mathbf{X}^{(j)}$ by using the list scores, i.e., $\{s^{(1,j)}, \ldots, s^{(N,j)} \}$ which bases on the Eqs. (\ref{eq:score_miae_ij}) and (\ref{eq:score-MIVAE-xij}). The final $\emph{AUC}$ is obtained by taking the maximum value of the $\emph{AUC}$ of all sub-encoders. The second way is called MIAEAD-sum and MIVAE-sum which is measured by using the list scores based on Eqs. (\ref{eq:score_miae_X}) and (\ref{eq:score-MIVAE-xi}). As shown in Table \ref{tab:max-vs-sum-results}, the $\emph{AUC}$ obtained by MIAEAD-max and MIVAE-max is significantly greater than that of MIAEAD-sum and MIVAE-sum. For example, MIAEAD-sum, MIAEAD-max, MIVAE-sum, and MIVAE-max achieve $0.715$, $0.886$, $0.797$, and $0.883$, respectively, in terms of average $\emph{AUC}$ across eight datasets. These results indicate that MIAEAD-max and MIVAE-max, which use feature subsets, can detect anomalies more effectively than MIAEAD-sum and MIVAE-sum, which use the entire feature space.

Third, we examine the performance of MIAEAD and MIVAE based on the number of sub-encoders used, as shown in Fig. \ref{fig:results-No-branches-used}. The results indicate that when the number of sub-encoders is small, the $\emph{AUC}$ obtained by MIAEAD and MIVAE is low. However, the $\emph{AUC}$ increases as the number of sub-encoders grows. For example, in Fig. \ref{fig:results-No-branches-used} (a) for MIAEAD and Fig. \ref{fig:results-No-branches-used} (b) for MIVAE, the $\emph{AUC}$ for the M3, M5, M6, and M8 datasets increases when the number of sub-encoders exceeds 15. These findings suggest that using feature subsets yields a higher $\emph{AUC}$ compared to using the entire feature set. 
Overall, the results help explain why MIAEAD and MIVAE, which use feature subsets to identify anomalies, can outperform other methods that use the entire feature space, as shown in Tables \ref{tab:experimental_results_01} and \ref{tab:experimental_results_02}.

\begin{table}[t]
	\caption{$\emph{AUC}$ obtained by MIAEAD and MIVAE by all sub-encoders and selecting the maximum value of sub-encoders.}
	\label{tab:max-vs-sum-results}
	\setlength\tabcolsep{1.5pt}
	\centering
	%\small
	\scriptsize
	%\footnotesize	
	\begin{tabular}{|c|c|c|c|c|c|c|c|c|c|c|}
		\hline
		Methods     & Types & M1 & M2 & M3 & M4 & M5 & M6 & M7 & M8 & \textbf{Average} \\ \hline \hline
		\multirow{2}{*}{MIAEAD} & sum   & \textbf{0.927} & 0.947& 0.764& 0.811& 0.538& 0.531& 0.715& 0.484& 0.715\\ \cline{2-11} 
		& max   & 0.921& \textbf{0.951} & \textbf{0.803} & \textbf{0.878} & \textbf{0.831} & \textbf{0.702} & \textbf{0.987} & \textbf{0.857} & \textbf{0.866}   \\ \hline \hline
		\multirow{2}{*}{MIVAE}  & sum   & \textbf{0.960} & \textbf{0.955} & 0.808& \textbf{0.902} & 0.473& 0.670& \textbf{0.990} & 0.616& 0.797\\ \cline{2-11} 
		& max   & 0.947& 0.953& \textbf{0.817} & 0.888& \textbf{0.898} & \textbf{0.755} & 0.988& \textbf{0.821} & \textbf{0.883}   \\ \hline
	\end{tabular}
\end{table}

\subsection{Influence of Ratio of Anomalies by Benign Samples}
\label{label:influence_ratio_anomaly_benign}

\begin{table*}[t]
	\centering
	\begin{center} \hspace*{-0.4cm} 
		\begin{tabular}{m{0em}cc} 
			&\begin{subfigure}{0.4\textwidth}\centering\includegraphics[width=\linewidth]{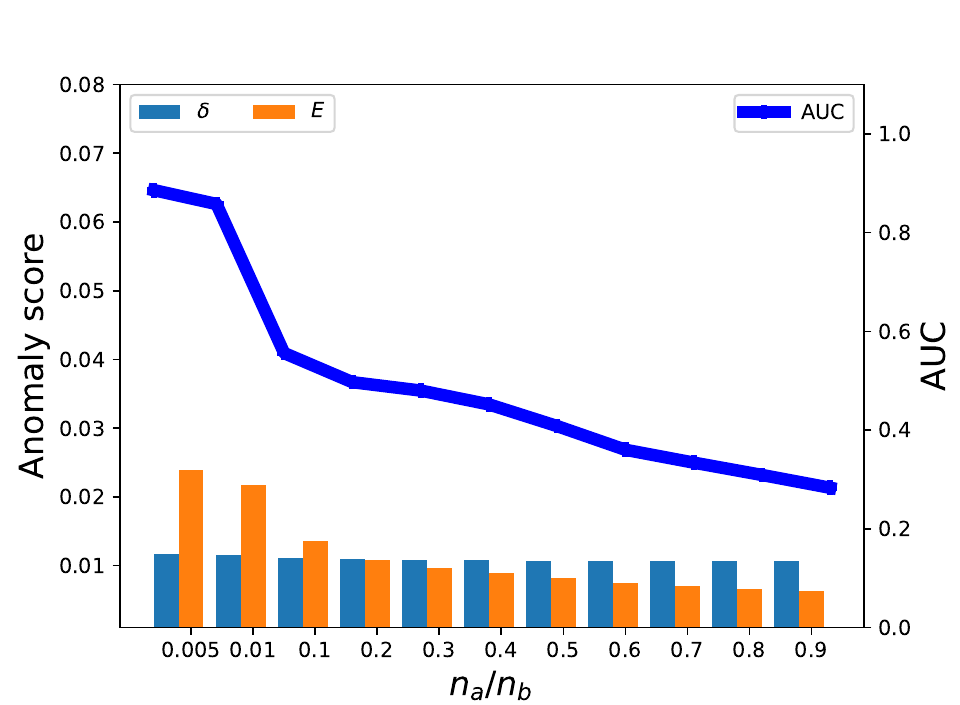}\end{subfigure}
			&\begin{subfigure}{0.4\textwidth}\centering\includegraphics[width=\linewidth]{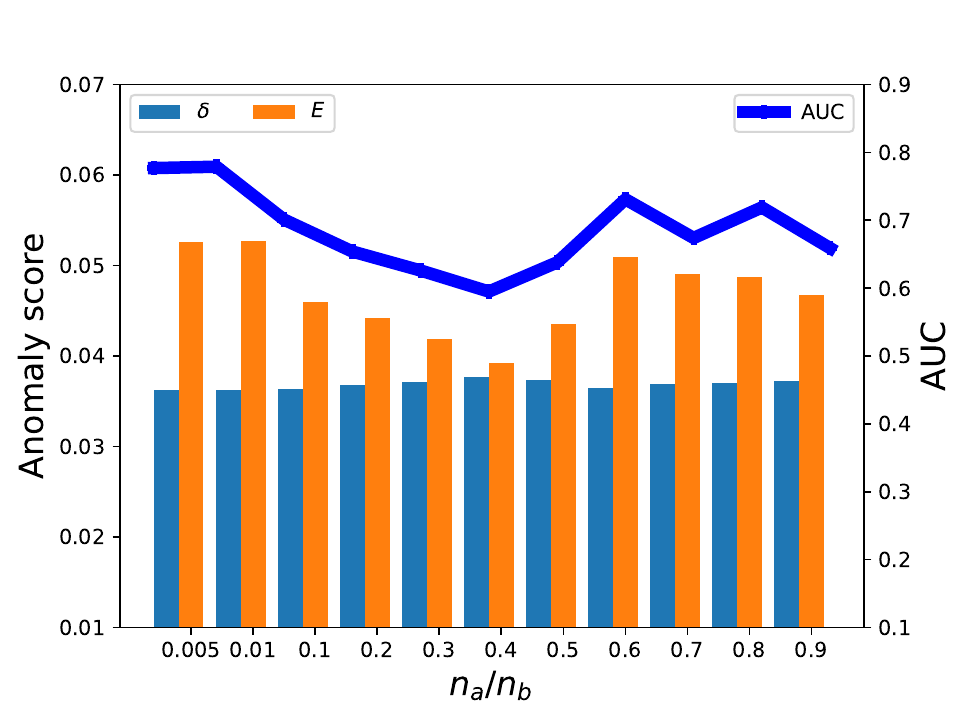}\end{subfigure} \\
            &(a) AEAD &(b) VAEAD \\ 
			&\begin{subfigure}{0.4\textwidth}\centering\includegraphics[width=\linewidth]{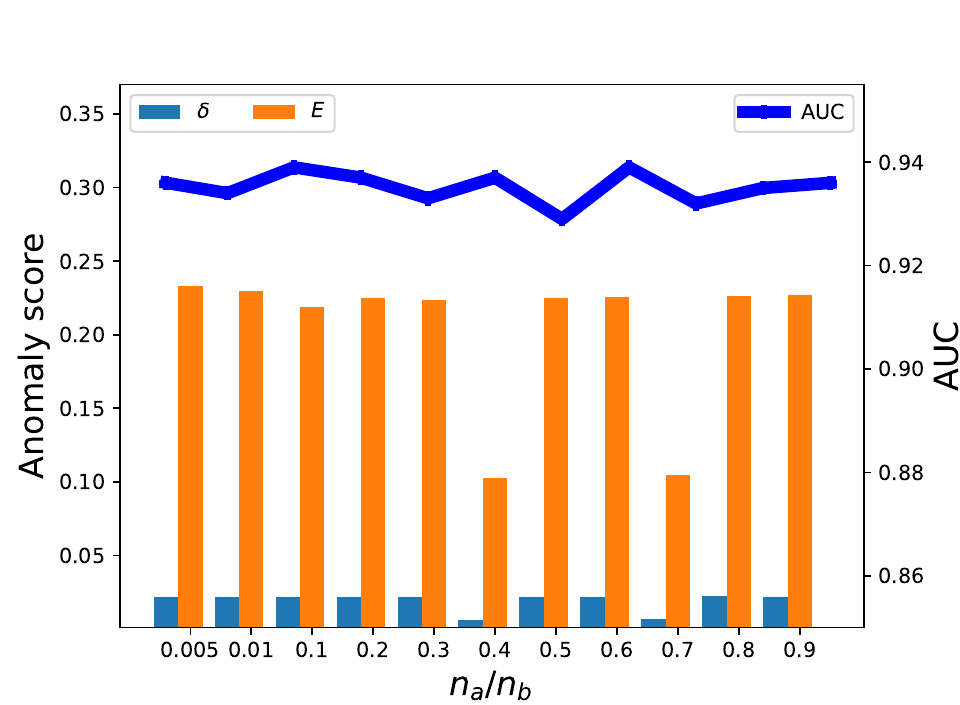}\end{subfigure}
			&\begin{subfigure}{0.4\textwidth}\centering\includegraphics[width=\linewidth]{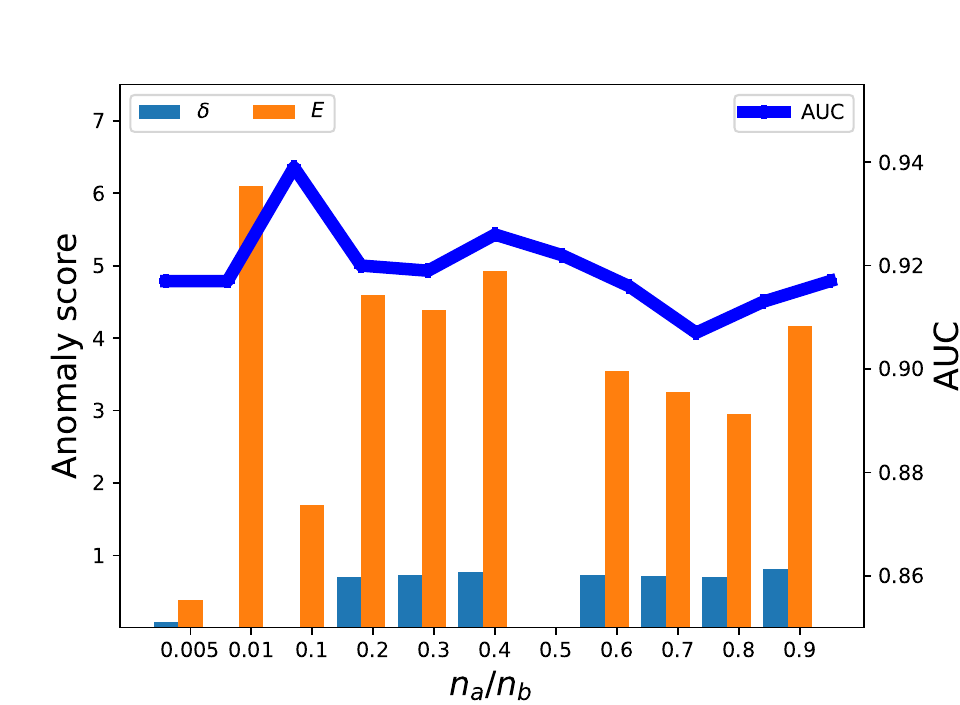}\end{subfigure}
			\\
			&(c) MIAEAD &(d) MIVAE  \\
			\\
		\end{tabular}
	\end{center} \vspace*{-0.5 cm}
	\captionof{figure}{ Results of AEAD, MIAEAD, and MIVAE based on the ratio of anomalies by the normal samples on the M5 dataset.}
	\label{fig:results-ratio-anomalies}
\end{table*}

We report the AUC obtained by anomaly detection methods as the ratio of anomalies to benign samples increases. We select the ratio of anomalies by benign samples $\frac{n_a}{n_b}$ by the list $\{0.005, 0.01, 0.1, 0.2, 0.3, 0.4, 0.5, 0.6, 0.7, 0.8, 0.9\}$. The experiments are tested on the M5 dataset with two methods, i.e., AE, MIAEAD, VAEAD, and MIVAE. The number of sub-encoders is 25. There are two scores, $\delta$ and $E$ are used. $\delta$ is an average anomaly score of benign samples, whilst $E$ is an average score of abnormal samples. Note that, for MIAEAD and MIVAE, $\delta$ and $E$ are calculated by sub-datasets.
First, as observed in Fig. \ref{fig:results-ratio-anomalies} (a), $\emph{AUC}$ obtained by AEAD decreases when the ratio $\frac{n_a}{n_b}$ increases. In addition, the values of $\delta$ are nearly unchanged, whilst the values of $E$ decline the same that of the $\emph{AUC}$. Interestingly, when the $\frac{n_a}{n_b} < 0.2$, the values of $\delta$ are greater than that of $E$. In contrast, $\delta < E$ if $\frac{n_a}{n_b} \geq 0.3$. This means that the AEAD only shows a high $\emph{AUC}$ with an assumption of a low ratio of anomalies by normal samples.

Second, we report the performance of VAEAD in Fig. \ref{fig:results-ratio-anomalies} (b). The $\emph{AUC}$ obtained by VAEAD shows a decline as the ratio of $\frac{n_a}{n_b}$ increases. Additionally, VAEAD achieves a higher $\emph{AUC}$ than AEAD. This is because $\delta$ is consistently lower than $E$, demonstrating the superior performance of VAEAD in modelling the distribution of normal data in the latent space to better identify anomalies.
Third, as shown in Fig. \ref{fig:results-ratio-anomalies} (c) and \ref{fig:results-ratio-anomalies}.d, the $\emph{AUC}$ obtained by MIAEAD and MIVAE fluctuates around $0.93$ and $0.92$, respectively. Additionally, the values of $E$ are consistently much greater than those of $\delta$. This indicates that the performance of MIAEAD and MIVAE is largely unaffected by the increasing ratio of anomalies to normal samples in the dataset.

To justify the influence of the ratio of anomalies to normal samples on the $\emph{AUC}$ obtained by AEAD, VAEAD, MIAEAD, and MIVAE, we propose a metric $\nu = \frac{E}{\delta}$ for $\delta > 0$. $\nu \gg 1$ indicates that the anomaly score can significantly distinguish anomalies from normal data, whereas $\nu \sim 1$ and $\nu < 1$ suggest that the anomaly score struggles to differentiate anomalies from normal samples. As shown in Table \ref{tab:explain_results_04}, the value of $\nu$ obtained by MIAEAD and MIVAE is significantly greater than $1$ as the ratio of anomalies to normal samples ($\frac{n_a}{n_b}$) increases. In contrast, the value of $\nu$ obtained by VAEAD is only slightly greater than $1$. For AEAD, $\nu < 1$ when $\frac{n_a}{n_b} \geq 0.2$. These results indicate that using feature subsets helps generate reconstruction errors for anomalies that are significantly higher than those for normal samples.

Overall, the results demonstrate that conventional methods, including AEAD and VAEAD, which use the entire feature space, may achieve lower $\emph{AUC}$ as the number of anomalies increases. In contrast, MIAEAD and MIVAE exhibit superior performance as the ratio of anomalies to normal samples changes, since they use feature subsets to identify anomalies.

%%%%%%%%%%%%%%%%%%%%%%%%%%%%%%%%%%%%
\begin{table}[t]
	\caption{Influence of $\nu :=\frac{E}{\delta}$ by the ratio of anomalies by the normal samples.}
	\label{tab:explain_results_04}
	\setlength\tabcolsep{1.5pt}
	\centering
	%\small
	\scriptsize
	%\footnotesize	
	\begin{tabular}{|c|c|c|c|c|c|c|c|c|c|c|c|c|}
		\hline
		$\frac{n_a}{n_b}$ & 0.005 & 0.01  & 0.1   & 0.2  & 0.3  & 0.4  & 0.5   & 0.6  & 0.7  & 0.8  & 0.9  & Average \\ \hline
		AEAD& 2.04  & 1.87  & 1.22  & 0.99 & 0.9  & 0.83 & 0.77  & 0.7  & 0.66 & 0.62 & 0.59 & 1.02    \\ \hline
		VAEAD & 1.45  & 1.45  & 1.26  & 1.2  & 1.13 & 1.04 & 1.16  & 1.4  & 1.33 & 1.32 & 1.25 & 1.27    \\ \hline
		MIAEAD& 3.35  & 3.31  & 3.17  & 3.21 & 3.19 & 3.19 & 3.17  & 3.19 & 3.16 & 3.16 & 3.18 & 3.21    \\ \hline
		MIVAE & 4.72  & 55.43 & 14.57 & 6.54 & 6.03 & 6.37 & 14.56 & 4.83 & 4.54 & 4.2  & 5.17 & 11.54   \\ \hline
	\end{tabular}
	
\end{table}

%%%%%%%%%%%%%%%%%%%%%%%%%%%%%%%%%%%%

\begin{figure}[t]
	\vspace*{0ex}
	\centering
	\includegraphics[width=0.4\textwidth]
	{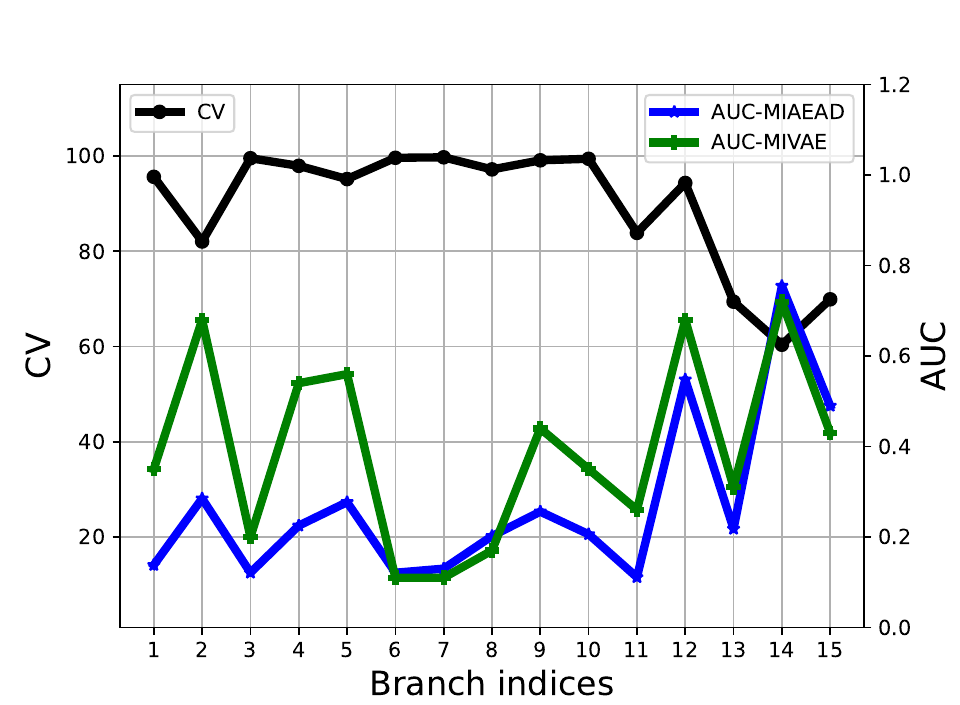}
	\caption{Influence of heterogeneity score, i.e., $\emph{CV}$, on $\emph{AUC}$. The number of sub-encoders used is 15, and the M5 dataset is used.}
	\label{fig:MIAEAD_CV_score}
\end{figure}

\subsection{Influence of Heterogeneity Score on Anomaly Detection Performance.}

We evaluate the influence of the heterogeneity score of the dataset, i.e., $\emph{CV}$, on the performance of the anomaly detection model, as observed in Fig. \ref{fig:MIAEAD_CV_score}. The $\emph{CV}$ and $\emph{AUC}$ are obtained from 15 feature subsets of the M5 dataset. We can draw four interesting results. First, the $\emph{CV}$ obtained by 15 feature subsets are different. This indicates that different feature subsets are heterogeneous, and features in a feature subset are also heterogeneous. Second, $\emph{AUC}$ obtained by MIAEAD and MIVAE has the same trend.
Third, the $\emph{CV}$ and $\emph{AUC}$ are likely inverse. For example, for sub-encoder numbers, i.e., 6, 7, and 8, the values of $\emph{CV}$ are high compared to the low values of $\emph{AUC}$. Fourth, the $\emph{AUC}$ obtained by MIVAE is greater than that of MIAEAD across most of the branch indices, as MIVAE can learn the distribution of normal data in the latent space. 
Overall, the result shows that features in a dataset are heterogeneous. Therefore, MIAEAD and MIVAE can detect anomalies in each feature subset instead of using all features to reduce the influence of heterogeneity of data.

\subsection{Tuning Hyper-parameters of MIVAE}
\begin{figure}[t]
	\vspace*{0ex}
	\centering \includegraphics[width=0.5\textwidth]
	{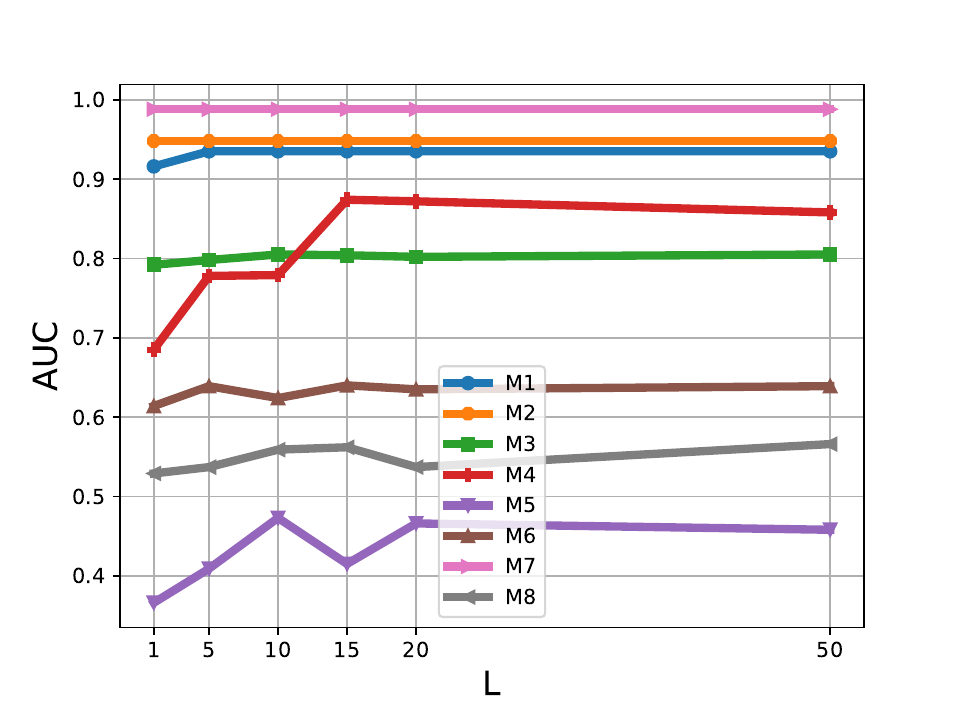}
	\caption{AUC obtained by MIVAE based on the hyper-parameter $L$.}
	\label{fig:MIVAE-hyper-parameter-L}
\end{figure}

\begin{table}[t]
	\centering
	\begin{tabular}{c} 
		\begin{subfigure}{0.5\textwidth}
			\centering
   \includegraphics[width=1.0\linewidth, height=0.3\linewidth]{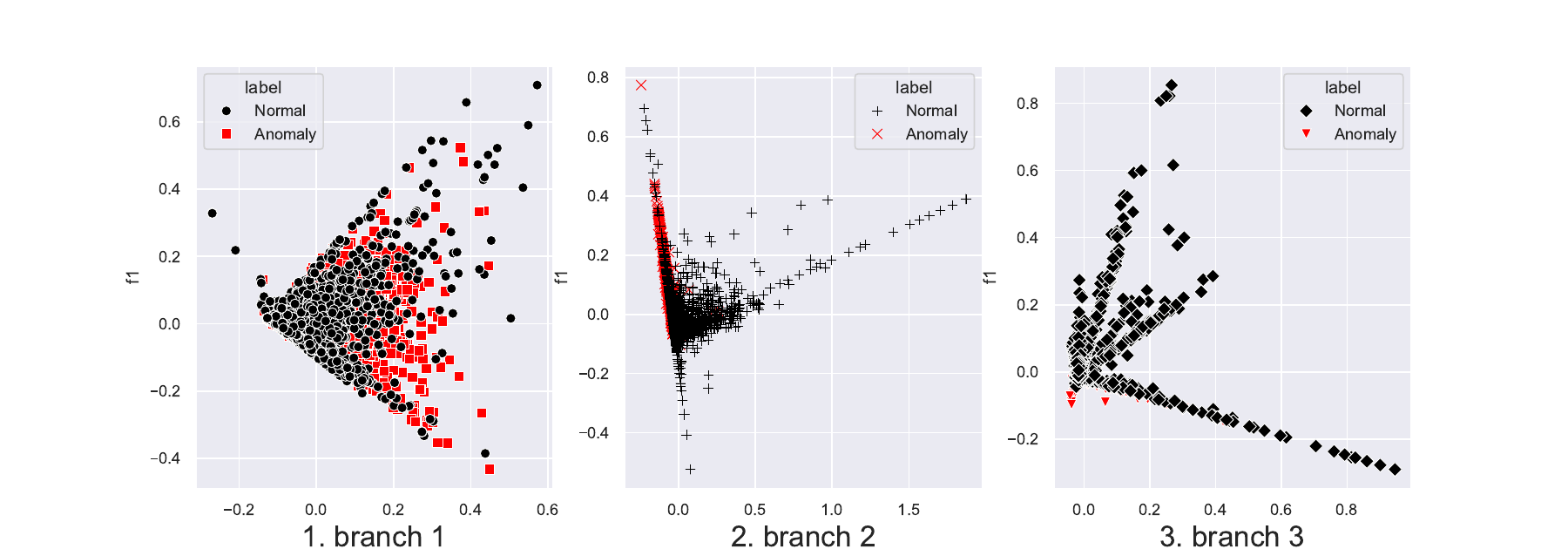}
			\caption{Original sub-datasets, i.e., $\textbf{X}^{(1)}$, $\textbf{X}^{(2)}$, and $\textbf{X}^{(3)}$.}
		\end{subfigure} \\
		\begin{subfigure}{0.5\textwidth}
			\centering
			\includegraphics[width=1.0\linewidth, height=0.3\linewidth]{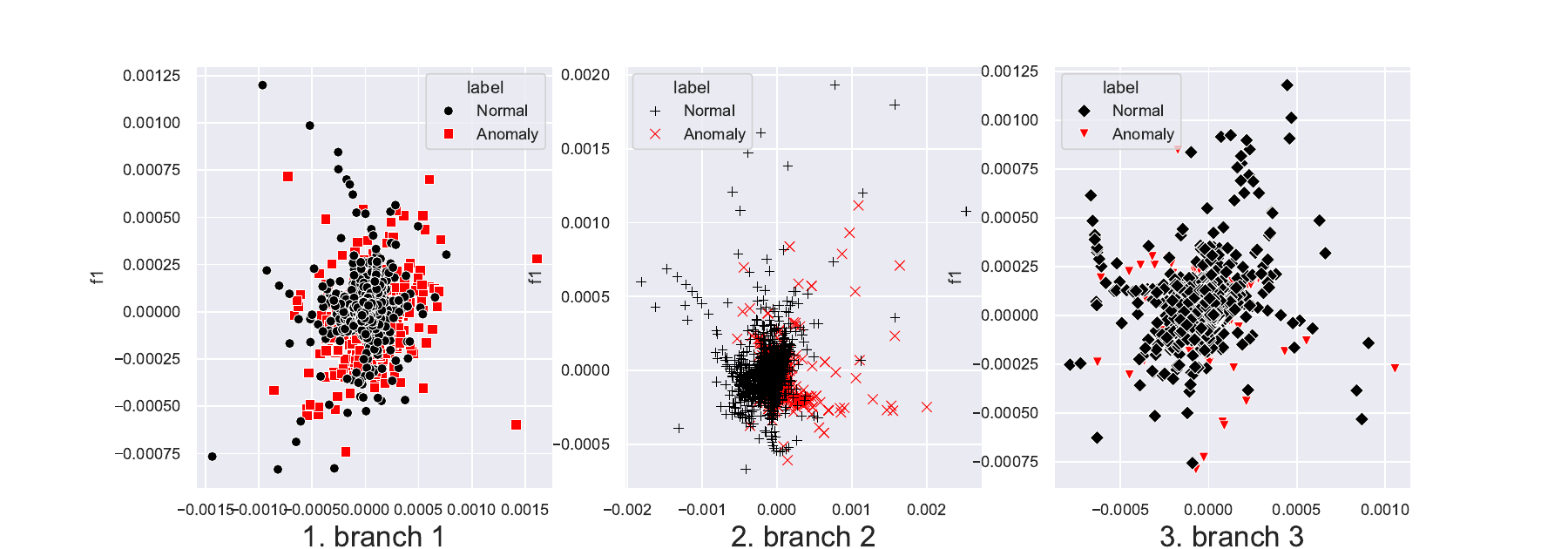}
			\caption{Data samples in latent space, i.e., $\textbf{e}^{(i,1)}$, $\textbf{e}^{(i,2)}$, and $\textbf{e}^{(i,3)}$, for $i=\{1, \ldots , N\}$.}
		\end{subfigure}\\
		\begin{subfigure}{0.5\textwidth}
			\centering
			\includegraphics[width=0.6\linewidth, height=0.3\linewidth]{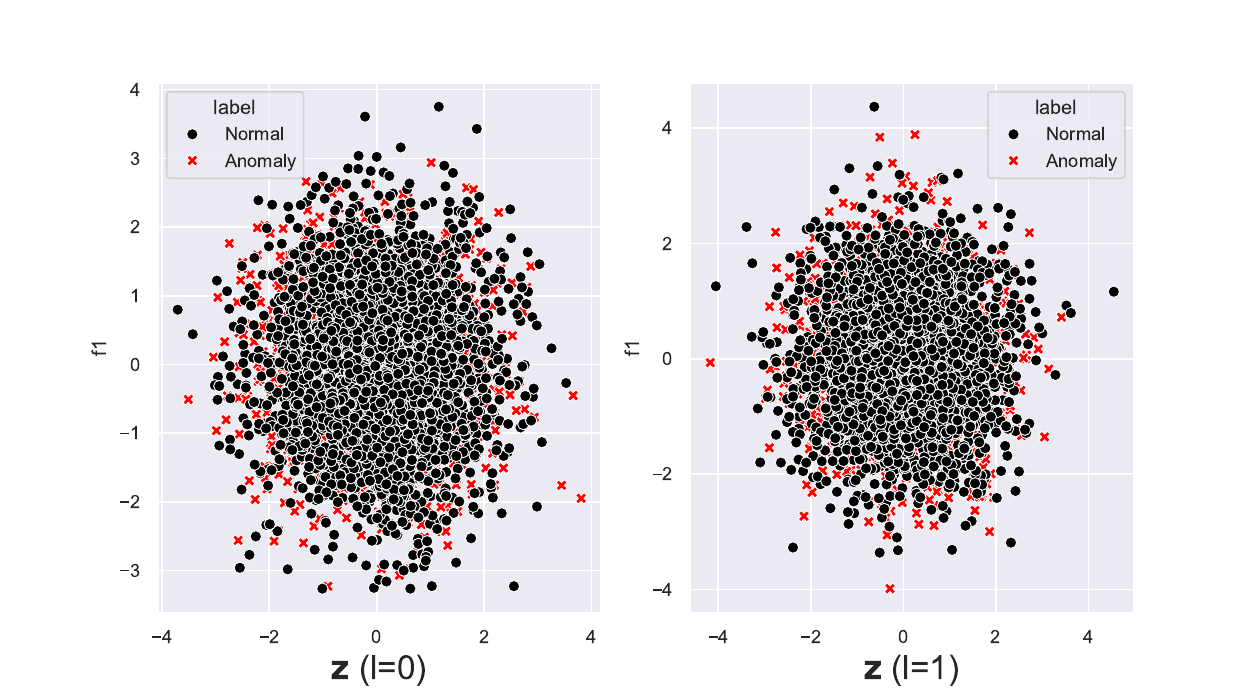}
			\caption{Generated data samples $\textbf{z}^{(i,l)}$ in the latent space for $l=\{1, 2\}$.}
		\end{subfigure}\\
		\begin{subfigure}{0.5\textwidth}
			\centering
			\includegraphics[width=1.0\linewidth, height=0.3\linewidth]{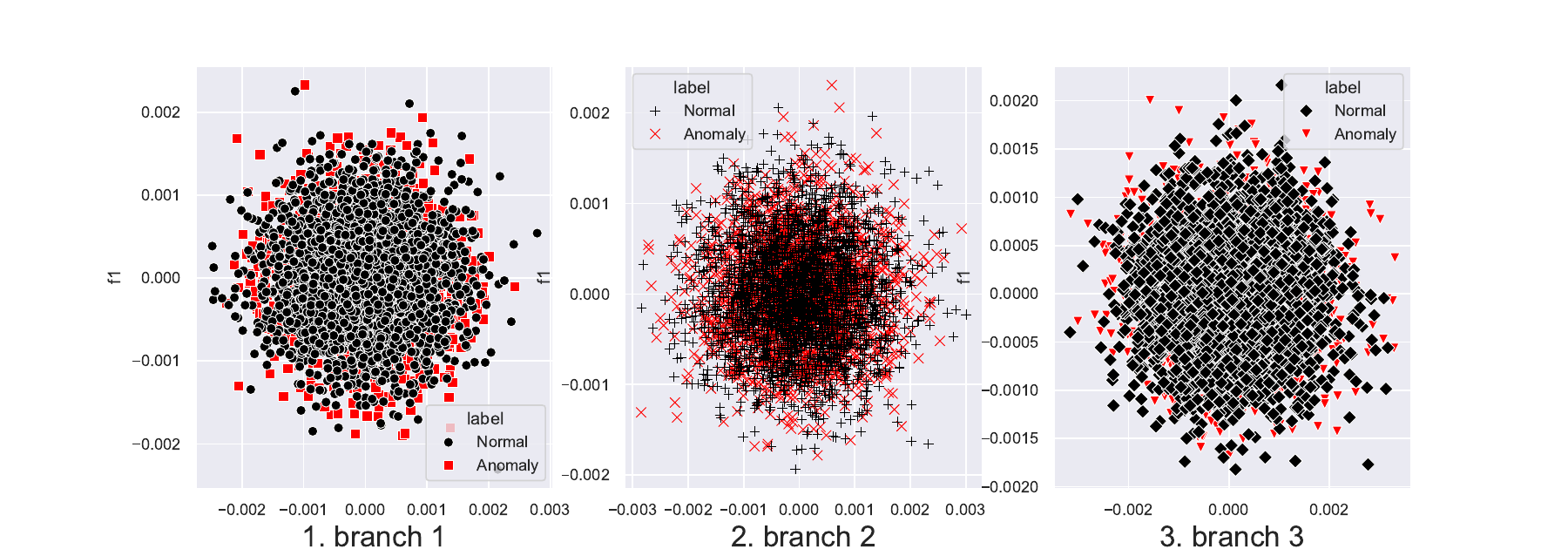}
			\caption{Reconstructed data samples $\hat{\textbf{x}}^{(i,j,l)}$ for $l=1$.}
		\end{subfigure}\\
		\begin{subfigure}{0.5\textwidth}
			\centering
			\includegraphics[width=1.0\linewidth, height=0.3\linewidth]{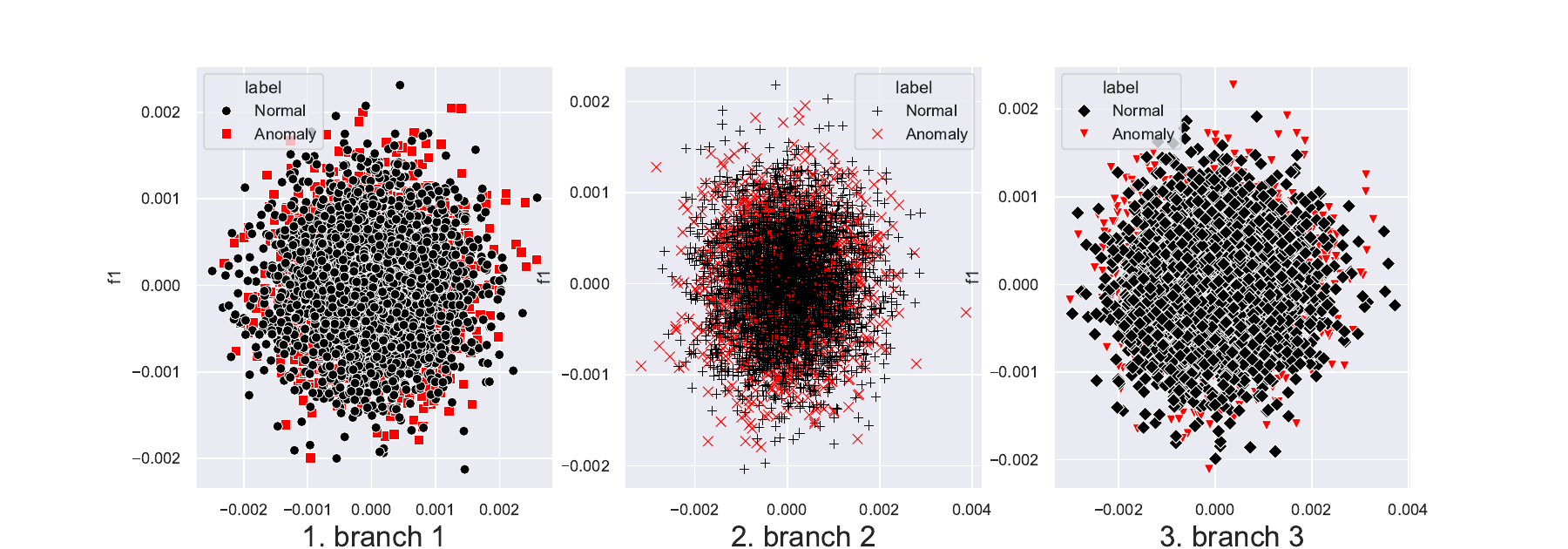}
			\caption{Reconstructed data samples $\hat{\textbf{x}}^{(i,j,l)}$ for $l=2$.}
		\end{subfigure}\\
	\end{tabular}
	%\vspace*{-0.3 cm}
	\captionof{figure}{Uncovering MIVAE through data simulation on the M8 dataset.}
	\label{fig:MIVAE-data-drawn}
\end{table}

We present the performance of MIVAE on the hyper-parameter $L$, which is the number of samples drawn in the latent space, as observed in Fig. \ref{fig:MIVAE-hyper-parameter-L}. The \emph{AUC} obtained by MIVAE is low for $L=1$, but stabilizes when $L \geq 10$. For example, the \emph{AUC} remains unchanged at about $0.94$ when $L \geq 5$ on the M1 dataset. This result indicates the effectiveness of modelling the distribution of normal data in the latent space, where anomalies may deviate from normal samples.

\subsection{Data Simulation of MIVAE}
Fig. \ref{fig:MIVAE-data-drawn} uncovers the data simulation of MIVAE through the M8 dataset. The dataset M8 includes three sub-datasets: $\mathbf{X}^{(1)}$, $\mathbf{X}^{(2)}$, and $\mathbf{X}^{(3)}$, which correspond to three sub-encoders of MIVAE. As observed in Fig. \ref{fig:MIVAE-data-drawn} (a).1, anomalies tend to distinguish from the normal samples, while both anomalies and normal samples overlap in Figs. \ref{fig:MIVAE-data-drawn} (a).2 and (a).3. The input data are transferred to the latent space $\mathbf{e}^{(1)}$, $\mathbf{e}^{(2)}$, and $\mathbf{e}^{(3)}$ to uncover latent features, and anomalies likely distinguish from normal samples in three branches in Fig. \ref{fig:MIVAE-data-drawn} (b). Next, the data samples $\mathbf{z}^{(i,l)}$ are shown in Fig. \ref{fig:MIVAE-data-drawn} (c), where anomalies may deviate from the Gaussian distribution of the normal data. Subsequently, Figs. \ref{fig:MIVAE-data-drawn} (d) and (e) present reconstructed data $\hat{\mathbf{x}}^{(i,j,l)}$ at the output of the decoder of MIVAE. We observe that anomalies tend to sit at the border of the Gaussian distribution. The data simulation helps explain why MIVAE can differentiate anomalies from normal samples by using sub-encoders to identify anomalies through feature subsets.

\subsection{Convergence Analysis and Anomaly Detection Performance of MIVAE}

To verify the convergence of MIVAE, we show the KL-divergence and loss functions (reconstruction errors) for each sub-encoder when MIVAE runs for 300 epochs on the M7 dataset. As observed in Fig. \ref{fig:sub-losses} (a), the MIVAE model converges after approximately 20 epochs. However, the $\emph{AUC}$ obtained by Branch $2$ is $0.83$, which is significantly lower than $0.98$ of Branches $1$ and $3$, even though the loss function line graphs for Branch $2$ and Branch $3$ are nearly the same. This discrepancy arises because the loss function is measured by the average anomaly score of both normal samples and anomalies.
To further explain, we plot the average anomaly score of normal samples $\delta$ and anomalies $E$, as observed in Fig. \ref{fig:sub-losses} (b). The line graphs of $\delta$ and $E$ for Branch $2$ are nearly overlapped, implying little difference between them. In contrast, the values of $E$ for Branches $1$ and $3$ are greater than those of $\delta$, resulting in a higher $\emph{AUC}$. These results demonstrate the effectiveness of using feature subsets to identify anomalies. We discuss histogram of anomaly scores obtained by three branches, as illustrated in Fig. \ref{fig:sub-losses} (c). The mean of the anomaly score distribution for normal samples obtained from Branch $1$ and Branch $3$ is lower than that for anomalies, while the mean of the anomaly score distribution for normal samples and anomalies may be the same on Branch $2$. This results in a higher AUC obtained by Branch $1$ and Branch $3$ compared to Branch $2$.

\begin{table}[t] \vspace*{-0.22 cm}
	\centering
	\begin{tabular}{c}
		\begin{subfigure}{0.45\textwidth} 
        \centering
			\includegraphics[width=1.0\linewidth, height=0.5\linewidth]{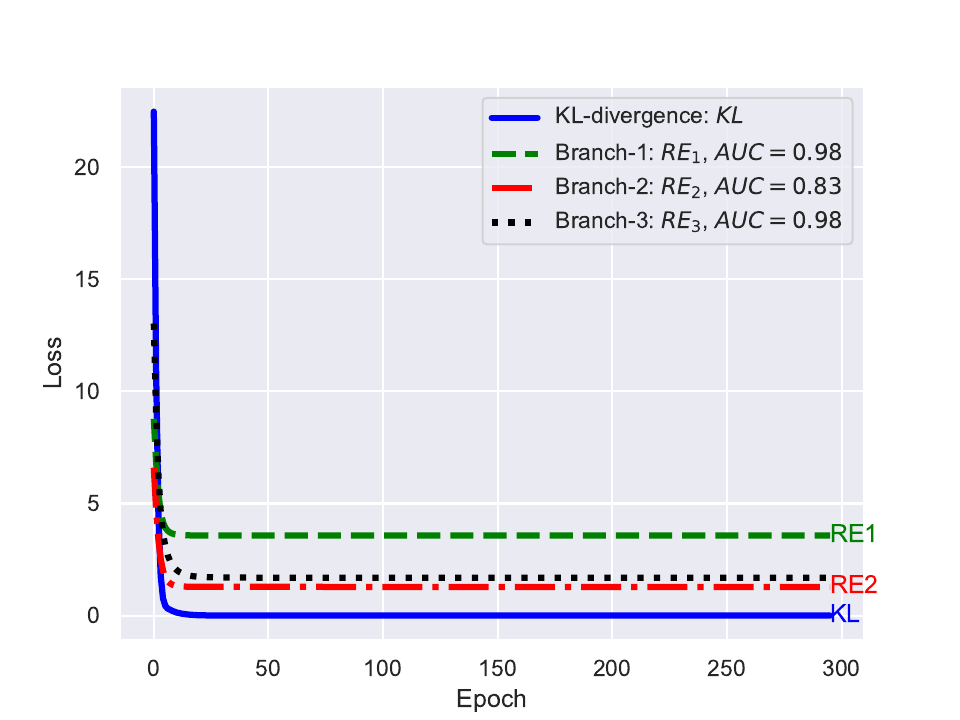} 
			\caption{Loss simulation.} 
		\end{subfigure}  \\
		\begin{subfigure}{0.45\textwidth}
        \centering
        %\vspace*{-0.15 cm}
            \includegraphics[width=1.0\linewidth, height=0.5\linewidth]{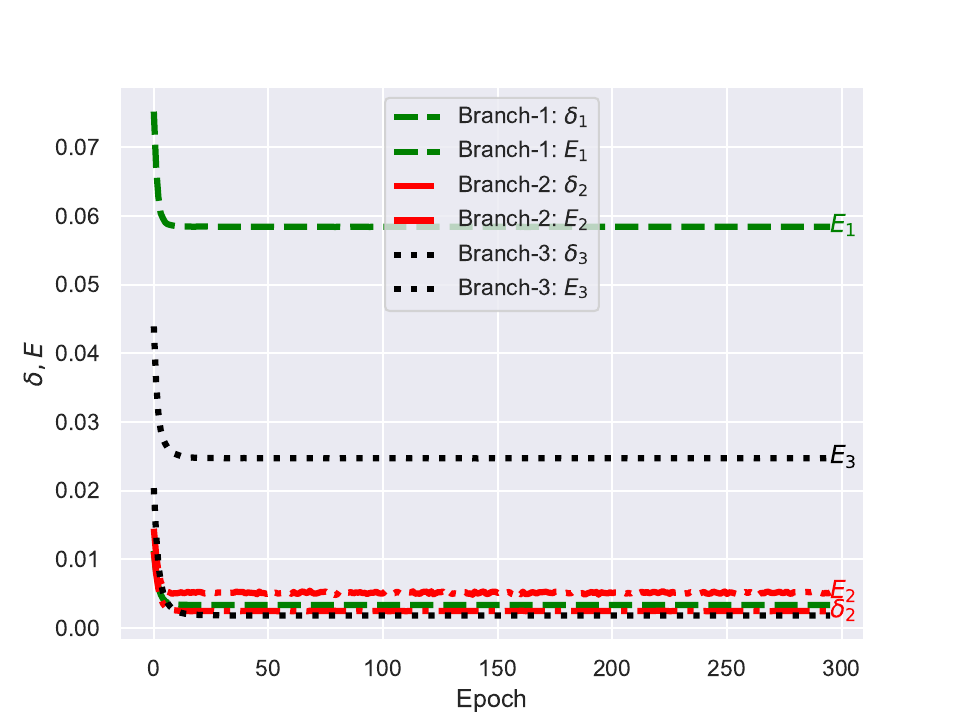}
			\caption{$E, \delta$ simulation.}
			\vspace{1.5ex} 
		\end{subfigure} \\ %\vspace*{-0.5 cm}
        \begin{subfigure}{0.5\textwidth}
        %\centering
        %\vspace*{-1.5 cm}
        \hspace*{-0.5 cm}
        \includegraphics[width=1.0\linewidth]{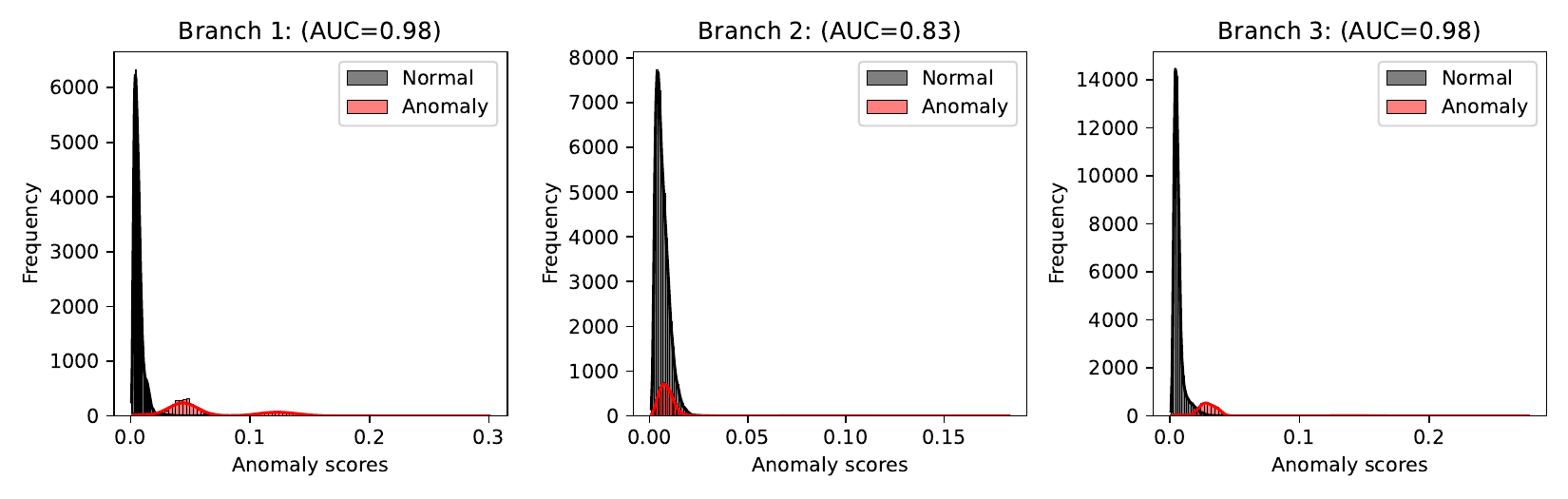}
			\caption{Histogram of anomaly score.}
			\vspace{1.5ex} 
		\end{subfigure} \\ %\vspace*{-0.5 cm}
		
	\end{tabular}
	\captionof{figure}{Loss simulation and anomaly score distribution of MIVAE.}
	
	\label{fig:sub-losses}
\end{table}

\section{Conclusions}
This paper proposed a novel deep learning model called MIAEAD for anomaly detection with non-IID data. MIAEAD assigned the anomaly score to each feature subset of a data sample that indicates the likelihood of being an anomaly. To do that, MIAEAD used the reconstruction error of its sub-encoder as the anomaly score of a feature subset. All sub-encoders were then simultaneously trained in an unsupervised learning technique to determine the anomaly scores of feature subsets. 
The final AUC obtained by MIAEAD was calculated for each sub-dataset, and the maximum AUC obtained among the sub-datasets was selected.
To leverage the modelling of the distribution of normal data to identify anomalies of the generative models, we then proposed a novel neural network architecture/model called MIVAE. MIVAE could process feature subsets through its sub-encoders before learning the distribution of normal data in the latent space. This allowed MIVAE to identify anomalies that significantly deviate from the learned distribution.

We theoretically proved that the difference in the average anomaly score between normal samples and anomalies obtained by the proposed MIVAE was greater than that of the VAEAD, resulting in a higher AUC for MIVAE. The MIAEAD and MIVAE models also used fewer parameters than the AEAD and VAEAD models, respectively. However, the training complexity of MIVAE was greater than that of VAEAD.
Extensive experiments on eight real-world anomaly datasets from different domains, e.g., health, finance, cybersecurity, and satellite imaging, demonstrated the superior performance of MIAEAD and MIVAE over conventional methods and the state-of-the-art unsupervised models, by up to $6\%$ in terms of AUC score.
Furthermore, experimental results showed that the AUC obtained by MIAEAD and MIVAE was mostly not impacted as the ratio of anomalies to normal samples in the dataset increased. We observed that MIAEAD and MIVAE had a high AUC when applied to feature subsets with low heterogeneity based on the coefficient of variation (CV) score. 

In the future, several directions can be explored. First, for datasets where the exact number of sub-datasets $M$ is unknown, one could develop a method to partition the dataset into $M$ sub-datasets to improve the AUC achieved by MIAEAD and MIVAE. Second, the MIAEAD and MIVAE can be effective for anomaly detection in image/video by using a convolutional neural network when a frame can be separated into different regions before inputting into MIAEAD and MIVAE \cite{sultani2018real}. Third, for time series data, one can divide the series into segments before feeding them into MIVAE, which can then be used to identify whether a specific segment is anomalous or normal \cite{10636792}. MIVAE can also recover missing information within a segment by generating data for the segment based on information from other segments in the series.

\section{Appendix}
\subsection{Proof of Theorem \ref{theo:theo_sub_score_ab}}
\label{l:proof-theorem1}

\begin{proof}
    The Theorem \ref{theo:theo_sub_score_ab} implies that the difference in the average anomaly score between normal samples and anomalies obtained by the proposed MIVAE is greater than that of the VAEAD, i.e., 
	$\Delta{s_{\emph{MIVAE}}} \geq \Delta{s_{\emph{VAEAD}}}$, resulting in a higher AUC for MIVAE. Based on the Hypothesis \ref{h:hypo_anomaly_score}, we accept that in anomaly detection using reconstruction error from VAEAD and MIVAE, anomalies exhibit higher reconstruction errors compared to normal samples, i.e., $ s_{(.)}(\mathbf{x}^{(i)}) < s_{(.)}(\mathbf{x}^{(k)}) $, where $ i = \{1, \ldots, n_b\} $ and $ k = \{n_b + 1, \ldots, n_b + n_a\} $ are the indices of the normal samples and anomalies in the dataset $ \mathbf{X}$, which can be separated by sub-datasets, i.e., $\mathbf{X} = \{\mathbf{X}^{(1)}, \mathbf{X}^{(2)}, \ldots, \mathbf{X}^{(M)} \}$. Assuming that dataset $\mathbf{X}$ has $M^{'}$ feature subsets being anomaly $j=\{1, \ldots, M^{'} \}$, where $ 1 \leq M^{'} \leq M$.

	  We prove that $\Delta^{(l)}{s_{\emph{MIVAE}}} \geq \Delta^{(l)}{s_{\emph{VAEAD}}}$ for $ l \in \{1, \ldots,L\}$. It means that the anomaly score of VAEAD and MIVAE is implemented by using one sampling iteration in their latent space, resulting in $\frac{1}{L} \sum_{l=1}^{L}\Delta^{(l)}{s_{\emph{MIVAE}}} \geq \frac{1}{L} \sum_{l=1}^{L} \Delta^{(l)}{s_{\emph{VAEAD}}}$. Therefore, we only need to prove $\Delta^{(l)}{s_{\emph{MIVAE}}} \geq \Delta^{(l)}{s_{\emph{VAEAD}}}$ for $L=1$.
  
  Based on Eqs. (\ref{eq:score_vae}) and (\ref{eq:delta}), we have $\Delta{s_{\emph{VAEAD}}} = s_{\emph{VAEAD}}(\mathbf{x}^{(k)}) - s_{\emph{VAEAD}}(\mathbf{x}^{(i)})$. 
	We have $s_{\emph{VAEAD}}(\mathbf{x}^{(i)}) = \delta$ and
 \begin{equation}
	\begin{aligned}
        s_{\emph{VAEAD}}(\mathbf{x}^{(k)}) &= \frac{1}{d} \big[ (d- \sum_{j=1}^{M^{'}}(d_j))  \delta + \sum_{j=1}^{M^{'}}(d_j  E) \big] \\ &= \delta + \sum_{j=1}^{M^{'}} \big[ \frac{d_j}{d}  (E - \delta) \big],
	\end{aligned}
\end{equation}
where $d_j$ is dimensionality of the sub-dataset $\mathbf{X}^{(j)}$. Therefore, 
 \begin{equation}
	\begin{aligned}
        \Delta{s_{\emph{VAEAD}}} =\delta + \sum_{j=1}^{M^{'}} \big[ \frac{d_j}{d}  (E - \delta) \big] - \delta = \sum_{j=1}^{M^{'}} \big[ \frac{d_j}{d}  (E - \delta) \big].
	\end{aligned}
\end{equation}

	For the proposed MIVAE, as observed in Eq. (\ref{eq:score-MIVAE-xi}), $s_{\emph{MIVAE}}(\mathbf{x}^{(i)}) = \sum_{j=1}^{M} (\beta^{(j)}  \delta)$ and 
 \begin{equation}
	\begin{aligned}
        s_{\emph{MIVAE}}(\mathbf{x}^{(k)}) = \sum_{j=1}^{M^{'}} (\beta^{(j)}  E ) + \sum_{j=M^{'}+1}^{M} (\beta^{(j)}  \delta)
	\end{aligned}
\end{equation} presents the average anomaly score for normal samples and anomalies, respectively. Therefore, 
\begin{equation}
	\begin{aligned}
        \Delta{s_{\emph{MIVAE}}} &= \sum_{j=1}^{M^{'}} (\beta^{(j)}  E ) + \sum_{j=M^{'}+1}^{M} (\beta^{(j)}  \delta) - \sum_{j=1}^{M} (\beta^{(j)}  \delta) \\ &= \sum_{j=1}^{M^{'}} (\beta^{(j)}  E )  + \sum_{j=M^{'}+1}^{M} (\beta^{(j)}  \delta) \\&- \sum_{j=M^{'}+1}^{M} (\beta^{(j)}  \delta) - \sum_{j=1}^{M^{'}} (\beta^{(j)}  \delta) \\ &= \sum_{j=1}^{M^{'}} \big[ \beta^{(j)}  (E - \delta)\big]
	\end{aligned}
\end{equation}

	To prove $\Delta{s_{\emph{MIVAE}}} \geq \Delta{s_{\emph{VAEAD}}}$ in Theorem \ref{theo:theo_sub_score_ab}, it is equivalent to proving 
 \begin{equation}
	\begin{aligned}
        \sum_{j=1}^{M^{'}} \big[ \beta^{(j)}  (E - \delta)\big] &\geq \sum_{j=1}^{M^{'}} \big[ \frac{d_j}{d}  (E - \delta) \big]  \\ \sum_{j=1}^{M^{'}} \beta^{(j)} &\geq \sum_{j=1}^{M^{'}} \frac{d_j}{d}
	\end{aligned}
\end{equation} with $E - \delta > 0$. 
	
\end{proof}

\begin{lemma}
\label{lem:lemma_betaj_greater_djd}
	$ \sum_{j=1}^{M^{'}} \beta^{(j)} \geq \sum_{j=1}^{M^{'}}  \frac{d_j}{d} $, where the $j^{th}$ sub-dataset contains anomalies, $j \in \{1, \ldots, M^{'} \}$.
\end{lemma}

\begin{proof}
	Based on Eq. (\ref{eq:beta-overline}), $\overline{\beta}$ is the average anomaly score for each feature for the entire dataset $\mathbf{X}$. $\beta^{(j)}$ presents the average anomaly score for each feature for the sub-dataset $\mathbf{X}^{(j)}$, as observed in Eq. (\ref{eq:beta-j}). We consider three cases as follows:
	\begin{itemize}
		\item If anomalies are presented only in one sub-dataset $ \mathbf{X}^{(j)} $, $M^{'}=1$, then $ d_j < d $ and $ \frac{d_j}{d} < 1 $. We have proven that $ \beta^{(j)} > 1 > \frac{d_j}{d} $, as observed in Remark \ref{remark_1}. Therefore, $ \sum_{j=1}^{M^{'}} \beta^{(j)} > \sum_{j=1}^{M^{'}}  \frac{d_j}{d}$.
		
		\item We consider a special case, where all feature subsets of anomalies are an anomaly, $M^{'} = M$. The traditional methods often consider if a data sample is an anomaly, its all features are anomalies. We have proven that $\beta^{(j)} = 1 > \frac{d_j}{d} $, as observed in Remark \ref{remark_2}. Therefore, $ \sum_{j=1}^{M^{'}} \beta^{(j)} > \sum_{j=1}^{M^{'}}  \frac{d_j}{d}$. 
		
		\item For generality, if anomalies present in $M^{'}$ sub-datasets with $1 < M^{'} < M$. The data sample $\mathbf{x}^{(i)}$ may contain many feature subsets which are anomalies, as observed in Fig. \ref{fig:sub-datasets}(b). We have proven generally that $\sum_{j=1}^{M^{'}} \beta^{(j)} \geq \sum_{j=1}^{M^{'}}  \frac{d_j}{d}$, as observed in Remark \ref{remark_3}. The equal sign happens when $M=1$. 
	\end{itemize}
\end{proof}

\begin{remark}
	\label{remark_1}
    Assume that dataset $\mathbf{X}$ has $M$ sub-datasets. There are  $M^{'}=1$ sub-dataset $\mathbf{X}^{(j)}$ which contains anomalies. We prove that $\beta^{(j)} > 1$.
\end{remark}
\begin{proof}
	The average anomaly scores for each feature of normal samples and anomalies are $\delta$ and $E$, respectively. The number of anomalies in the sub-dataset $\mathbf{X}^{(j)}$ is $n_a$. We have $\overline{\beta} = \frac{1}{M  N} \big[ n_a  E + (M N -n_a)  \delta \big]$, as observed in Eq. (\ref{eq:beta-overline}). Therefore in Eq. (\ref{eq:beta-j}), 
  \begin{equation}
	\begin{aligned}
        \beta^{(j)} = \frac{1}{N  \overline{\beta}}  \big[ n_a  E + (N-n_a)  \delta \big]  = M  \frac{n_a  E + (N-n_a)  \delta}{n_a  E + (M  N - n_a)  \delta}.
	\end{aligned}
\end{equation}
	We have 
\begin{equation}
\begin{aligned}
    \beta^{(j)} &> 1 \\ M  \frac{n_a  E + (N - n_a)  \delta}{n_a  E + (M  N - n_a)  \delta} &> 1 \\ 
     M  n_a  E + M  (N - n_a)  \delta  &> n_a  E + (M  N - n_a)  \delta \\ 
     M  n_a  E - n_a  E   &> (M  N - n_a - M  N + M  n_a)  \delta \\ 
     (M - 1)  n_a  E &> (M - 1)  n_a  \delta \\ 
     E &> \delta.
\end{aligned}
\end{equation}
 This holds because for any dataset $\mathbf{X}$ having $d>1$, we can separate it into sub-datasets, leading to $M>1$. In addition, $E > \delta$ follows by the Hypothesis \ref{h:hypo_anomaly_score}.
\end{proof}

\begin{remark}
	\label{remark_2}
    Assume that dataset $\mathbf{X}$ has $M$ sub-datasets and $M^{'}=M$ sub-datasets contains anomalies. For every anomaly samples $\mathbf{x}^{(k)}$, all feature subsets are anomaly. We prove that $\beta^{(j)} = 1$.
\end{remark}

\begin{proof}
	We have  
\begin{equation}
	\begin{aligned}
        \overline{\beta} = \frac{1}{M  N}  \big[M  n_a  E + M (N -n_a)  \delta \big] = \frac{1}{N}  \big[ n_a  E + (N -n_a)  \delta \big] 
	\end{aligned}
\end{equation}
 and $\beta^{(j)} = \frac{1}{\overline{\beta}  N} \big[ n_a  E + (N-n_a)  \delta \big]= \frac{\overline{\beta}}{\overline{\beta}} =1$. 
\end{proof}

\begin{remark}
	\label{remark_3}
    Assume that dataset $\mathbf{X}$ has $M$ sub-datasets and $M^{'}$ sub-datasets contains anomalies, where $M \geq M^{'} \geq 1$. The numbers of anomalies in $M^{'}$ sub-datasets are $n_{a_1}, \ldots, n_{a_{M^{'}}}$, where $a_j \in \{1,\ldots, M\}$ and $ 1 \leq n_{a_j} \leq n_a$. We prove that $ \sum_{j=1}^{M^{'}} \beta^{(j)}  \geq \sum_{j=1}^{M^{'}} \frac{d_j}{d}  $.
\end{remark}

\begin{proof}
	We have 
 \begin{equation}
	\begin{aligned}
        \overline{\beta} = \frac{1}{M  N}  \big[ \sum_{j=1}^{M^{'}} (n_{a_j}  E) + \big(M N - \sum_{j=1}^{M^{'}}n_{a_j} \big)  \delta \big] 
	\end{aligned}
\end{equation}

\begin{flalign}
\begin{aligned}
    \beta^{(j)} &= \frac{1}{\overline{\beta}  N} \big[ n_{a_j}  E + ( N - n_{a_j})  \delta \big] \\ 
    &= M  \frac{n_{a_j}  E + ( N - n_{a_j})  \delta}{ \sum_{j=1}^{M^{'}} (n_{a_j}  E) + \big(M  N - \sum_{j=1}^{M^{'}} n_{a_j} \big)  \delta} \\ 
    &= \frac{M  n_{a_j}  (E - \delta) + M  N  \delta}{\sum_{j=1}^{M^{'}} (n_{a_j}  (E - \delta)) + M  N  \delta} \\ 
    &= \frac{M  n_{a_j} + \frac{M  N  \delta}{E - \delta}}{\sum_{j=1}^{M^{'}} n_{a_j} + \frac{M  N  \delta}{E - \delta}}  
    = \frac{M  n_{a_j} + C}{\sum_{j=1}^{M^{'}} n_{a_j} + C},
\end{aligned}
\end{flalign}
where $C := \frac{M N  \delta}{E-\delta} > 0$. Next, 
 \begin{equation}
	\begin{aligned}
        \sum_{j=1}^{M^{'}} \beta^{j} &= \sum_{j=1}^{M^{'}} \bigg[ \frac{M  n_{a_j} + C }{\sum_{j=1}^{M^{'}} n_{a_j} + C }  \bigg]= \frac{\sum_{j=1}^{M^{'}} \big[ M  n_{a_j} + C \big]}{\sum_{j=1}^{M^{'}} n_{a_j} + C }  \\ &= \frac{M \sum_{j=1}^{M^{'}} n_{a_j} + M^{'}  C}{\sum_{j=1}^{M^{'}} n_{a_j} + C} = \frac{M A + M^{'}  C}{ A + C},
	\end{aligned}
\end{equation}
 where $A :=\sum_{j=1}^{M^{'}} n_{a_j} > 0$. For $M \geq M^{'} \geq 1$, we have $\sum_{j=1}^{M^{'}} \beta^{j} \geq 1$ because 
 \begin{equation}
	\begin{aligned}
         M A + M^{'}  C &\geq A+C  \\  (M-1)  A + (M^{'}-1) C &\geq 0; \\  
         \forall A >0; C>0; M \geq M^{'} &\geq 1 .
	\end{aligned}
\end{equation}
	We have $\sum_{j=1}^{M^{'}} \frac{d_j}{d} \leq \sum_{j=1}^{M} \frac{d_j}{d}  = \frac{d}{d} = 1$. Finally, $ \sum_{j=1}^{M^{'}} \beta^{(j)}  \geq 1 \geq \sum_{j=1}^{M^{'}} \frac{d_j}{d} $.
\end{proof}

\subsection{Proof of Lemma \ref{lem:lemma_parameter_encoder}}
\label{l:lamda1}
\begin{proof}
We need to show that $p_t(\emph{VAEAD}) >  p_t(\emph{MIVAE})$, where
  \begin{equation*}
	\begin{aligned}
        p_t(\emph{VAEAD})= (\alpha_t  d + 1)  \alpha_{t+1}  d = \alpha_t  \alpha_{t+1}  d^2 + \alpha_{t+1} d,
	\end{aligned}
\end{equation*}
\begin{equation}
	\begin{aligned}
        p_t(\emph{MIVAE}) &= \sum_{j=1}^{M} \big((\alpha_t  d_j + 1)  \alpha_{t+1}  d_j \big) \\ &= \sum_{j=1}^{M} \big(\alpha_t \alpha_{t+1} d_j^2 \big) + \sum_{j=1}^{M} \big(\alpha_{t+1} d_j \big) \\ &= \alpha_t  \alpha_{t+1} \sum_{j=1}^{M}d_{j}^{2} + \alpha_{t+1}  d
	\end{aligned}
\end{equation} because $d = \sum_{j=1}^{M} d_j$, $\forall d_{j} >0, j=\{1, \ldots, M\}$.   
We aim to calculate
\begin{equation}
	\begin{aligned}
        p_t(\emph{VAEAD}) -  p_t(\emph{MIVAE})  &= \alpha_t \alpha_{t+1} \big( d^2 - \sum_{j=1}^{M}d_{j}^{2} \big) \\ &= \alpha_t \alpha_{t+1} \big( (\sum_{j=1}^{M} d_j)^2 - \sum_{j=1}^{M}d_{j}^{2} \big) \\ &= 2 \alpha_t \alpha_{t+1} \sum^{}_{1 \leq k < j \leq M} (d_k  d_j) > 0 ; \\ & \forall d_{j}, d_{k} >0, \alpha_t, \alpha_{t+1} >0.
	\end{aligned}
\end{equation} 	
Therefore, 
$p_t(\emph{VAEAD})   > p_t(\emph{MIVAE})$.

\end{proof}

%%%%%%%%%%%%%%%%%%%%%%%%%%%%%%%%%%%%%%%%%%%%%%%%%%%%%%%%%%%%%%%%%%%%%%%%%%%%%%%%%%%%%
%\section*{Acknowledgement} 
%“This research was supported in part by the UTS-VNU Joint Technology and Innovation Research Centre and the Australian Research Council under the DECRA project DE210100651.”
\bibliographystyle{IEEEtran}
\bibliography{IEEEabrv,library}{}

%\vfill

%\appendices
%\section{Proof of the First Zonklar Equation}
%Appendix one text goes here.

% you can choose not to have a title for an appendix
% if you want by leaving the argument blank
%\section{}
%Appendix two text goes here.

% that's all folks
\end{document}